\documentclass[preprint,12pt]{elsarticle}
\usepackage[utf8]{inputenc}
\usepackage{textcomp}
\usepackage{graphicx}
\usepackage{amsmath}

\usepackage[version=4]{mhchem}
\usepackage[group-separator={,},group-minimum-digits={3}]{siunitx}
\usepackage{longtable,tabularx}
\usepackage{algorithm}
\usepackage{algpseudocode}
\usepackage{amsfonts}
\usepackage{bm}
\usepackage{booktabs}
\usepackage{multirow}
\usepackage{gensymb}
\usepackage{url}
\usepackage{subfiles}
\usepackage{tikz}
\usetikzlibrary{fit}
\usetikzlibrary{shapes,arrows,decorations.pathmorphing,shadows}
\usetikzlibrary{positioning}
\usetikzlibrary{patterns}
\usepackage{nomencl}
\makenomenclature
\usepackage{etoolbox}
\renewcommand\nomgroup[1]{%
  \item[\bfseries
  \ifstrequal{#1}{G}{Terminology}{%
  \ifstrequal{#1}{O}{Symbols}{}}%
]}
\journal{Acta Astronautica}

\begin{document}

\begin{frontmatter}

\title{A Lattice-based Method for Optimization in Continuous Spaces with Genetic Algorithms}
% \author{Cameron Harris, Kevin Schroeder, Jonathan Black}

% \author{Cameron D. Harris \footnote{Graduate Student, Kevin T. Crofton Department of Aerospace and Ocean Engineering, AIAA Student Member} and Kevin B. Schroeder \footnote{Associate Director, National Security Institute, AIAA Member}}
% \affil{Virginia Tech, Blacksburg, VA, 24061}
% \author{Jonathan Black\footnote{Professor, Kevin T. Crofton Department of Aerospace and Ocean Engineering, AIAA Associate Fellow}}
% \affil{Virginia Tech, Blacksburg, VA, 24061}

\author{Cameron D. Harris} %% Author name
\author{Kevin B. Schroeder} %% Author name
\author{Jonathan Black} %% Author name

%% Author affiliation
\affiliation{organization={Virginia Tech},%Department and Organization
            % addressline={},
            city={Blacksburg},
            postcode={24061},
            state={Virginia},
            country={USA}}

% \date{July 2024}

\begin{abstract}

  This work presents a novel lattice-based methodology for incorporating multidimensional constraints into continuous decision variables within a genetic algorithm (GA) framework.
  The proposed approach consolidates established transcription techniques for crossover of continuous decision variables, aiming to leverage domain knowledge and guide the search process towards feasible regions of the design space.
  This work offers a robust and general-purpose lattice-based GA that is applicable to a broad range of optimization problems.
  Monte Carlo analysis demonstrates that lattice-based methods find solutions two orders of magnitude closer to optima in fewer generations.
  The effectiveness of the lattice-based approach is showcased through two illustrative multi-objective design problems: (1) optimal telescope placement for astrophotography and (2) optimal design of a satellite constellation for maximizing ground station access.
  The optimal telescope placement example shows that lattice-based methods converge to the Pareto front in 15\% fewer generations than traditional methods.
  The orbit design example shows that lattice-based methods discover an order of magnitude more Pareto-optimal solutions than traditional methods in a highly constrained design space.
  Overall, the results show that the lattice-based method exhibits enhanced exploration capabilities, traversing the solution space more comprehensively and achieving faster convergence compared to conventional GAs.

\end{abstract}

%%Graphical abstract
% \begin{graphicalabstract}
% %\includegraphics{grabs}
% \end{graphicalabstract}

%%Research highlights
% \begin{highlights}
% \item Research highlight 1
% \item Research highlight 2
% \end{highlights}

%% Keywords
\begin{keyword}
%% keywords here, in the form: keyword \sep keyword

genetic algorithm \sep constrained optimization \sep orbit design

%% PACS codes here, in the form: \PACS code \sep code

%% MSC codes here, in the form: \MSC code \sep code
%% or \MSC[2008] code \sep code (2000 is the default)

\end{keyword}

\end{frontmatter}

% \section*{Nomenclature}

% \noindent(Nomenclature entries should have the units identified)

% {\renewcommand\arraystretch{1.0}
% \noindent\begin{longtable*}{@{}l @{\quad=\quad} l@{}}

%   \multicolumn{2}{@{}l}{Genetic algorithm terminology}\\
%   generation & an iteration of the genetic algorithm \\
%   population & the set of solutions in a generation \\
%   chromosome & a single solution in a population \\
%   gene & the positional elements that comprise a chromosome \\
%   allele & a value realized by a gene \\

%   \multicolumn{2}{@{}l}{Symbols}\\
%   $X$ & a chromosome \\
%   $\chi$ & a gene \\
%   $\alpha$ & an allele \\
%   $\Theta$ & a set of linked genes $\theta$ \\
%   $\theta$ & a gene within a linked set of genes $\Theta$ \\
%   $\phi$ & latitude \\
%   $\Phi$ & set of latitudes $\phi$ \\
%   $\lambda$ & longitude \\
%   $\Lambda$ & set of longitudes $\lambda$ \\
%   $\Gamma$ & scaling factors for Gaussian lattice \\
%   $n_p$ & node number (lattice configuration parameter) \\
%   $n_q$ & quantile number (Gaussian lattice configuration parameter) \\
%   $a$ & satellite orbit semi-major axis, \unit{km} \\
%   $i$ & satellite orbit inclination \\
%   $\Omega$ & satellite orbit longitude of the ascending node \\

% \end{longtable*}}

\nomenclature[G]{\(generation\)}{an iteration of the genetic algorithm}
\nomenclature[G]{\(population\)}{the set of solutions in a generation}
\nomenclature[G]{\(chromosome\)}{a single solution in a population}
\nomenclature[G]{\(gene\)}{the positional elements that comprise a chromosome}
\nomenclature[G]{\(allele\)}{a value realized by a gene}

\nomenclature[O]{\(\Omega\)}{a chromosome}
\nomenclature[O]{\(\omega\)}{a gene}
\nomenclature[O]{\(\alpha\)}{an allele}
\nomenclature[O]{\(\Theta\)}{a set of linked genes}
\nomenclature[O]{\(\theta\)}{a gene within a linked set}
\nomenclature[O]{\(\phi\)}{latitude}
\nomenclature[O]{\(\Phi\)}{set of latitudes $\phi$}
\nomenclature[O]{\(\lambda\)}{longitude}
\nomenclature[O]{\(\Lambda\)}{set of longitudes $\lambda$}
\nomenclature[O]{\(\Gamma\)}{scaling factors for Gaussian lattice}
\nomenclature[O]{\(n_p\)}{node number (lattice configuration parameter)}
\nomenclature[O]{\(n_q\)}{quantile number (Gaussian lattice configuration parameter)}
\nomenclature[O]{\(a\)}{satellite orbit semi-major axis, \unit{km}}
\nomenclature[O]{\(i\)}{satellite orbit inclination}
\nomenclature[O]{\(\Omega\)}{satellite orbit longitude of the ascending node}

\printnomenclature

%% Begin file imports for the sections

% Include beginning sections
\section{Introduction}\label{sec:introduction}

% brief description of evolutionary algorithms
It is difficult to pose many applied engineering design problems tractably or efficiently for modern optimization methods because real-world models often have complex, non-convex design spaces.
Metaheuristic optimization approaches show considerable promise in this area, as these methods operate on the decision variables instead of the design space.
Evolutionary computation, and the genetic algorithm (GA) in particular, has become a popular tool for engineering design problems~\cite{man1996genetic, colombi2018multi, arias2012multiobjective, wagner2020genetic, wagner2021distributed, lee2023designing, yokoyama2005modified, zheng2017swarm}.
However, practical engineering design problems may have requirements that limit the actionable design space, which is a challenge point for GAs.

A GA is an optimization technique that mimics biological evolution to find fit solutions for given objectives.
Genetic algorithms treat individual solutions as members of a population.
Members that perform well against objectives are selected to reproduce to create the next generation of the population, intending to combine advantageous features from high performing individuals to produce even better performing individuals in the proceeding generation.
Continued improvement ultimately leads to individuals with unsurpassable fitness in one or several objectives.
Additionally, mutations to individuals are applied stochastically to encourage population diversity and search of previously unexplored areas of the design space.

Fundamentally, all genetic algorithms follow the same procedure \cite{mitchell1998introduction}:

\begin{enumerate}
    \item Generate a random population of candidate solutions
    \item Evaluate the fitness of solutions
    \item Select the high performing candidate
    \item Recombine selected candidates to produce new solutions
    \item Apply random mutations to new solutions
    \item Repeat steps 2-6 until stopping criteria is satisfied
\end{enumerate}

% explain why evolutionary algorithms are useful
The GA encodes each solution as a sequence of values, referred to as a chromosome.
Each gene in the chromosome is a design choice, taking on a value from the range of admissable values in the design space.
The specific value taken by a gene is called an allele.
As noted above, GAs operate on the population instead of the search space, facilitating the robustness for sufficiently complex engineering problems.
% Because the GA is not a gradient descent algorithm, it is capable of searching over both discrete and continuous decision variables.
Because the GA uses a population-based search strategy, it is well-suited for problems with discrete, continuous, or mixed variable types.

In constrained optimization, decision variables must satisfy specified conditions to be considered permissible solutions to the problem at hand;
these conditions are referred to as constraints.
A solution that satisfies all constraints in an optimization problem is referred to as a ``feasible'' solution.
Traditional GAs may inadvertently explore infeasible regions of the solution space, generating solutions that violate one or more constraints.
Such solutions are of no practical value in design problems, as they do not represent viable designs.
Therefore, the ability to effectively handle constraints is a critical feature for GAs to be considered a useful tool in design optimization.
By incorporating constraint handling, GAs can be adapted to efficiently search for feasible solutions within the constrained solution space, leading to more meaningful and actionable results.

Several distinct methods of constraint handling in GAs have been researched, varying in application, complexity, and generality~\cite{coello2002theoretical, michalewicz1995genetic}.
The literary history of constraint handling in GAs is broad and somewhat fragmented, as many techniques have been developed in isolation to address particular use-cases.
However, in general, constraint handling approaches fall roughly into three groups: penalty methods, repair methods, and specialized operators.

The most popular method of constraint handling in genetic algorithms is to apply a penalty to infeasible solutions~\cite{coello2002theoretical, michalewicz1995genetic, homaifar1994constrained, kuri2002penalty}.
% References \cite{coello2002theoretical, michalewicz1995genetic} summarizes several methods for constraint handling in GAs.
The techniques described by \cite{michalewicz1995genetic} suggest methods of levying penalties in an effort to guide search towards feasible regions of the search space.
Nonetheless, infeasible solutions still appear in the population, with the goal that sufficiently high penalty precludes the candidate from selection in future generations.

Empirically, penalty methods in stochastic optimization cannot guarantee both the prevention of infeasible solutions and algorithm convergence.
In other words, depending on the convergence criterion and configuration of the genetic algorithm, the solutions returned by the algorithm are not guaranteed to be feasible.
As discussed in \cite{davis1987genetic}, a large penalty for constraint violations may result in premature convergence or stagnation when a feasible solution is found.
Conversely, a low penalty may result in infeasible solutions being returned, as they may have higher fitness than the feasible solutions.
Indeed, the results of \cite{michalewicz1995genetic} show that, for some methods, infeasible solutions were present in the optimal set of solutions.
% In principle, penalty methods should incentivize search in regions where penalties are not present.
Therefore, tuning of the penalty factors may be required to achieve desired performance, and the correct tuning of the parameters is not known a-priori.

In some cases, gene repair is a viable strategy to ensure a candidate satisfies constraints~\cite{coello2002theoretical, ponsich2008constraint}.
With repair, the allele in violation of a constraint is manually overwritten before the objective functions are evaluated.
Repair is generally defined as the minimum possible modification to the allele(s) to satisfy constraints.
Gene repair requires an a-priori derived method, likely based on a problem-dependent approach, to find the nearest feasible solution.
While repair is effective when there exists a straightforward methodology, it cannot be used in all constrained applications.
Furthermore, depending on the repair strategy, repair may also lead to oversampling along the constraint boundaries in the design space.

Specialized operator approaches broadly encompass all algorithms that modify genetic operators from their traditional usage.
Many specialized operator methods have been developed to address constraint handling.
One example of an approach with specialized genetic operators is an agent-based genetic algorithm, which directs candidate solutions towards feasible regions through agent learning~\cite{barkat2008search}.
Agent-based approaches show promise, but introduce considerable complexity over a traditional genetic algorithm.
A second example is the GENOCOP algorithm, which is a problem-independent method for solving constrained optimization problems~\cite{michalewicz1996genocop}.
GENOCOP is a constraint consistent algorithm, where infeasible decision variables are never present in a genome.
The underlying methodology ensures solutions satisfy constraints by selecting variables with stricter constraints first and isolating the search spaces of the remaining decision variables to the feasible space.
However, GENOCOP is limited only to problems with linear constraints.

Importantly, constraints are not the only challenge facing GAs in the context of design.
In practical applications, decision variables may have interdependence, which is a phenomenon known as epistasis~\cite{davidor1990epistasis}.
An epistatic relationship between two genes means the value of one gene suppresses or enhances the expression of another gene.
Traditionally, genetic algorithms operate on genes individually and stochastically, so dependent relationships between genes are not considered.
However, it has been demonstrated that insensitivity to epistasis may considerably hinder performance~\cite{salomon1996re}.

Overall, there exists a need for an all-purpose, efficient GA that handles arbitrary constraints and is sensitive to epistatic genes.
The work presented in this article shows that established specialized operator methods can be synthesized into a robust approach suitable for constraint handling.
The approach is generalizable and operates well with linear and nonlinear constraints.
Two examples in different state space representations illustrate the method's general applicability.

The proceedings of this article are organized as follows.
Section \ref{sec:motivation} provides a motivating example with traditional methods.
Section \ref{sec:methodology} outlines the methodology of a new technique.
Section \ref{sec:results} analyzes the performance and applies the proposed technique to two sample problems.
Section \ref{sec:conclusion} provides some concluding remarks and outlines some potential future work.
% Appendix \ref{app:universities} contains some data used to produce the results in Section \ref{sec:conclusion}.

\section{Motivation}\label{sec:motivation}

% Traditional crossover techniques are not designed for constrained search spaces.
% A continuous search space can be reduced to a discrete search space.
% This limits the the available operations that can occur during crossover; specifically, a child must inherit an exact copy of alleles from its parents, rather than an average or other mixing operation.

% While discretizing the search space by itself presents a path forward for optimization, a discrete search space on its own lacks positional relationships between points.

To better demonstrate the limitations of traditional methods, a motivating highly constrained example is provided below.
Consider a geographic search problem, where a single location is designated as the optimum and solutions are constrained to land only.
Complex geography renders some regions infeasible for search, but feasible regions are continuous over their domain.
The decision variables that determine geographic location are the combination of latitude, $\phi$, and longitude, $\lambda$.

The latitude-longitude space lends itself well to a heuristic for gene repair.
Genes repair is known to be an effective constraint handling mechanism when a heuristic is available.
As described in Section \ref{sec:introduction}, gene repair overwrites infeasible alleles with feasible values.

In this motivating example, if a location violates a constraint, the alleles will be modified to the closest feasible latitude-longitude pair, where the closeness is measured by physical distance between the location and constraint boundary.
% This can be done using the $min$ function over the coastlines provided by the MATLAB Mapping Toolbox \cite{MATLAB_MAPPING_Toolbo
The constraint boundary is transcribed into a finite set of points, $\Phi^{boundary}, \Lambda^{boundary}$, to facilitate repair.
The nearest feasible latitude-longitude may then be found by the minimum arclength from the set of points that transcribe the boundary.
The MATLAB mapping toolbox provides built-in coastline boundary data \cite{MATLAB_MAPPING_Toolbox}.

The central angle between two latitude-longitude coordinate pairs is given by the haversine formula, as written in Eq. (\ref{eq:haversine_formula}).

\begin{equation} \label{eq:haversine_formula}
    \textrm{hav}\left(\phi_1, \lambda_1, \phi_2, \lambda_2\right) = \frac{1}{2}\left(\textrm{ver}\left(\phi_2 - \phi_1\right) + \cos\left(\phi_1\right)\cos\left(\phi_2\right)\textrm{ver}\left(\lambda_2 - \lambda_1\right)\right)
\end{equation}

The haversine is the literal concatenation of ``half the versine'', where the versine is defined in Eq. (\ref{eq:versine}).

\begin{equation} \label{eq:versine}
    \textrm{ver}\left(x\right) = 1 - \cos\left(x\right)
\end{equation}

With the ability to compute distances between geographical locations, it is possible to find the minimum distance the feasible region for repair.
The formula to repair an infeasible location, parameterized by coordinates $\left(\phi^{infeasible}, \lambda^{infeasible}\right)$, is provided in Eq. (\ref{eq:repair_latlon}).

\begin{equation} \label{eq:repair_latlon}
    \begin{gathered}
        i = \textrm{argmin}(\textrm{hav}(\Phi^{boundary}, \Lambda^{boundary}, \phi^{infeasible}, \lambda^{infeasible})) \\
        \phi^{repaired}, \lambda^{repaired} = \Phi^{boundary}_i, \Lambda^{boundary}_i
    \end{gathered}
\end{equation}

In Eq. (\ref{eq:repair_latlon}), the minimum distance is calculated between the infeasible coordinates $\left(\phi^{infeasible}, \lambda^{infeasible}\right)$ and the set of coordinates that define the constraint boundary.
The index of the closest point $i$ is found, such that $\left(\phi^{infeasible}, \lambda^{infeasible}\right)$ describes the closest point in the set of all points that transcribe the constraint boundary.
The infeasible coordinates $\left(\phi^{infeasible}, \lambda^{infeasible}\right)$ are then replaced by coordinates at $\left(\phi^{repaired}, \lambda^{repaired}\right)$, concluding the coordinate repair.

\begin{figure}[ht!]
    \centering
    \includegraphics[width=0.7\textwidth]{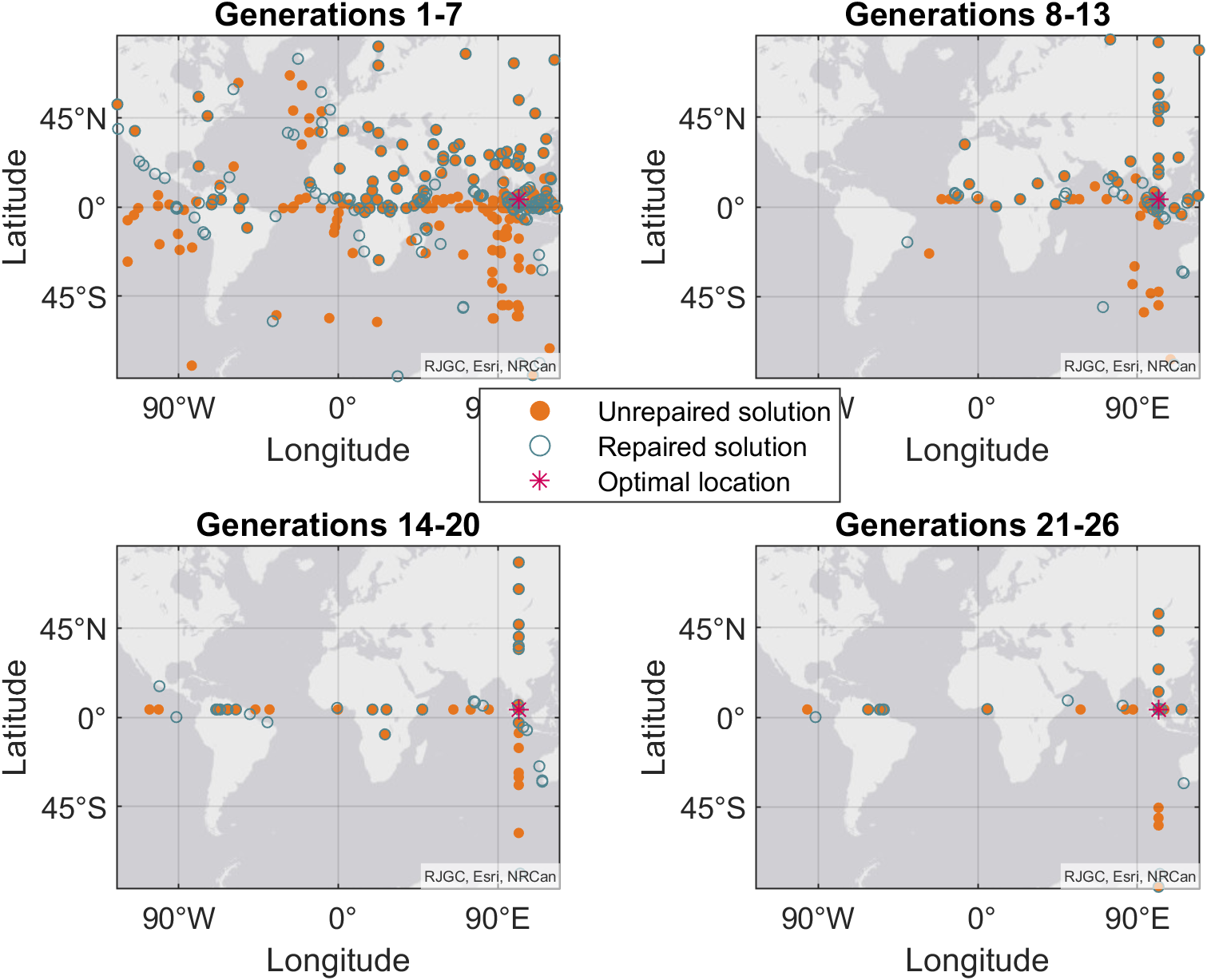}
    \caption{Example of population evolution with repair. The population diversity collapses quickly due to the independent treatment of the epistatic latitude and longitude genes.}
    \label{fig:repair_frames}
\end{figure}

Figure \ref{fig:repair_frames} shows a population's evolution when using a traditional genetic algorithm, where infeasible solutions are repaired.
As the generations advance, the population's diversity collapses around the solution with high fitness.
As traditional crossover and mutation operators modify genes individually, the distribution of un-repaired candidates is a cross pattern after five or so generations.
Once the candidates are repaired, the solutions heavily sample the boundaries of the constrained space nearest to the cross pattern.
The consequence is that proceeding global search -- driven by mutation -- is biased towards a fraction of the design space.
Population diversity is generally known to be an important condition for effective search~\cite{arias2012multiobjective, gupta2012overview}.
Research shows that homogeneity in a population reduces selection pressure, leading to stagnation and premature convergence~\cite{whitley2001overview}.

Figure \ref{fig:repair_frames} is a visual representation of the stagnation risk associated with constraint inconsistent genetic operators.
The design feature of interest is the location, which is described by the combination of latitude and longitude parameters.
After a few generations, a high fitness solution is found and quickly propagates through the population.
In this instance, the optimal location was near the boundary, which resulted in most offspring of the high fitness solution landing in an infeasible area.
The infeasible offspring were repaired to the same location, so local search stagnated and diversity was lost.

The diversity collapse is also partially driven by the independent treatment of latitude and longitude;
mutations to only one gene are not effective when the genes are interdependent.
An epistatic relationship exists between the latitude and longitude genes because the choice of the longitude gene significantly impacts the effectiveness of the choice of the latitude gene, and vice-versa.
Global search falters after a few generations because mutations to individual genes are not sensitive to epistasis.

\section{Methodology}\label{sec:methodology}

This work proposes a method of multi-dimensional search that prioritizes exploration of related parameters and directly incorporates constraints.
In a GA, search is performed by two evolutionary operations: crossover and mutation.
Crossover involves the recombination of genes between two parents to produce an offspring with traits from both parents, facilitating local search.
Mutation randomly resamples genes from anywhere in the admissable space, facilitating global search.

This work develops specialized crossover and mutation operators that generalize and expand on existing literature to handle constrained optimization problems.
Leung and Wang \cite{leung2001orthogonal} demonstrated a method of domain transcription for crossover of continuous decision variables in GAs.
The work systematizes the transcription process for multi-dimensional design spaces, so that the continuous space enclosed by two parents is discretized into a finite set of points.
The finite set of points, or nodes, generated from recombining two parents is referred to in this work as a lattice.
This work intends to expand on the contributions of \cite{leung2001orthogonal} by adapting the approach for constraint handling and generalizing the approach to handle alternative lattice structures (see Section \ref{subsec:lattice_based_crossover}).

Particularly, the approach presented in this work follows the principles of constraint consistency, where candidates that violate constraints cannot exist in the population~\cite{kowalczyk1997constraint}.
This work will show how constraints, if present, may be propagated into the genetic operators to prevent infeasible solutions from appearing in offspring, similar to GENOCOP~\cite{michalewicz1996genocop} but expanded to handle nonlinearities.

The genetic algorithm, NSGA-II~\cite{deb2002fast}, is of principle focus for this work, though it may be generalized to other variations of the GA.
The genetic operators of primary importance are crossover and mutation.
In this framework, implementation requires that crossover and mutation operations be configured for each decision variable.
Genes that are linked by epistasis or constraints must be jointly handled in crossover and mutation operations to ensure that the respective modifications do not produce an infeasible solution.

While many crossover operators exist, typically a single crossover operation is defined for all genes in a GA optimization program.
In this proposed lattice-based method, the crossover operator acts on each set of linked decision variables.
A book-keeping mechanism is required in the program to keep track of which genes are linked to each other.
If the parent chromosome is $X$, then a set of linked genes in $X$ will be denoted as $\Theta$, such that $\Theta \subseteq X$.
A set of linked genes, $\Theta \in \mathbb{R}^N$, is composed of $N$ genes, such that $\Theta = [\theta_1, \theta_2, ...,\theta_N]$.

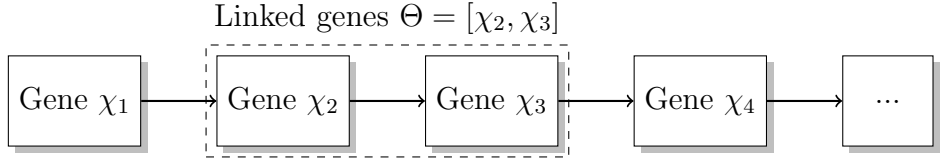
\begin{figure}[ht!]
    \centering
    \begin{tikzpicture}
        % Define gene styles
        \tikzstyle{gene} = [draw=black, fill=white, rectangle, minimum size=1.2cm, drop shadow];

        % Create chromosome
        \node (A) [gene] {Gene $\chi_1$};
        \node (B) [gene, right=1.0cm of A] {Gene $\chi_2$};
        \node (C) [gene, right=1.0cm of B] {Gene $\chi_3$};
        \node (D) [gene, right=1.0cm of C] {Gene $\chi_4$};
        \node (E) [gene, right=1.0cm of D] {...};

        % Create linked gene group
        \node [draw=black, dashed, fit=(B)(C), label=above:Linked genes {$\Theta = [\chi_2, \chi_3]$}, decorate] {};

        % Connect genes with arrows
        \draw[->, thick] (A) -- (B);
        \draw[->, thick] (B) -- (C);
        \draw[->, thick] (C) -- (D);
        \draw[->, thick] (D) -- (E);

    \end{tikzpicture}
    \caption{Example chromosome $X$ containing two linked genes, $\chi_2$ and $\chi_3$.}
    \label{fig:chromosome_example}
\end{figure}

For example, consider the arbitrary chromosome $X$ depicted in Fig. \ref{fig:chromosome_example}, where an arbitrary constraint exists such that the constraint is only a function of the second and third genes, $\chi_2$ and $\chi_3$.

The genes $\chi_2$ and $\chi_3$ are linked by the presence of the constraint.
The first (and only) set of linked genes $\left[\chi_2, \chi_3\right]$ is denoted as $\Theta$, such that

\begin{equation*}
    \Theta = \left[\chi_2, \chi_3\right] = \left[\theta_1, \theta_2\right]
\end{equation*}

Note that there may be multiple sets of linked genes in a given optimization problem.
Each set of linked genes $\Theta$ is the input to the respective crossover and mutation operators for $\Theta$.
The crossover and mutation operators are explained in more detail below.

% Propagating domain information into the crossover operator permits the crossover to restrict child genes to the feasible design space.
% Therefore, for every set of linked alleles $\Theta$, there must be a crossover operator defined and applied to $\Theta$.

% Typically a single crossover , there is literary precedent for varying the crossover operator on a per-gene basis.
% For instance, the satellite constellation design studies performed in \cite{wagner2020optimization} considers real-valued genomes containing continuous and discrete decision variables.
% The crossover operation for two continuous decision variables, say $\theta_1$ and $\theta_2$, is a weighted average such that $\theta_{1,new} = \theta_1 - r(\theta_1 - \theta_2)$, where $r \in [0, 1]$ is randomly sampled.
% However, the crossover operation for two discrete decision variables merely swaps the values, with no additional modification (i.e., $r=1$).

\subsection{Lattice-based crossover} \label{subsec:lattice_based_crossover}

The general algorithm for lattice-based crossover is (a) construct a lattice from the alleles of two parents, (b) randomly sample nodes from the lattice, (c) when a feasible node is found, return the node as the offspring.
At least one set of alleles from the parents is included in the lattice, so there is always at least one feasible node in the lattice.
Therefore, it is guaranteed that the lattice-based crossover returns a feasible solution.

The pseudocode framework for the lattice crossover operator is outlined in Algorithm \ref{alg:lattice_crossover}.
Denote the set of linked genes from a parent chromosome as $\Theta$, and the resulting offspring's alleles as $\Theta'$.
A general parameter $n_p$ is introduced, which determines the number of nodes in each level of the lattice.
An abstract function $\Call{ConstructLattice}$ is called, which produces the set of nodes that form the lattice.
There may be additional configuration parameters specific to an implementation of $\Call{ConstructLattice}$, which are indicated by the trailing ellipses in the inputs of Algorithm \ref{alg:lattice_crossover} and function signature of $\Call{ConstructLattice}$.
Algorithms \ref{alg:uniform_lattice} and \ref{alg:gauss_lattice} provide example implementations of $\Call{ConstructLattice}$.

\begin{algorithm}
    \hspace*{\algorithmicindent} \textbf{Input}: $\Theta^A$, $\Theta^B$, $n_p$, ... \Comment{additional configuration inputs may be needed}\\
    \hspace*{\algorithmicindent} \textbf{Output}: $\Theta^{A'}$
    \caption{Lattice Quantization}\label{alg:lattice_crossover}\
    \begin{algorithmic}
        % \State \Comment{Generate chromosome for first child from lattice centered on Parent A}
        % Step 1: Construct lattice based on alleles
        \State $Lattice \gets $ \Call{ConstructLattice}{$\Theta^A, \Theta^B$, $n_p$, ...}  \Comment{See Algorithms \ref{alg:uniform_lattice} and \ref{alg:gauss_lattice}}
        % \State $Lattice_A \gets $ \Call{ConstructLattice}{$chromosome_A, chromosome_B$}\Comment{See Algorithms \ref{alg:gauss_lattice} and \ref{alg:grid_lattice}}
        % Step 2: Check if each member of the lattice is valid
        % \State Step 2: Evaluate each node in the lattice for constraint violations. Remove nodes that violate constraints.
        % \State Randomly order lattice
        \For{$nodes \in Lattice$}  \Comment{Randomly order lattice}
            \If{$node$ satisfies constraints}
                % \State $ValidNodes \overset{+}{\leftarrow} node$
                \State $\Theta^{A'} \gets node$
                \State Break loop
            \EndIf
        \EndFor
        % Step 3: Sample from valid members in lattice
        % \State Step 3: Randomly select $\Theta'$ from the remaining set of nodes that satisfy all constraints.
        % \State $\Theta^{A'} \gets$ \Call{RandomSample}{$ValidNodes$}

        \State \Comment{Repeat steps with reversed inputs for $\Theta^{B'}$}

    \end{algorithmic}
\end{algorithm}

% \begin{figure}
%     \centering
%     \begin{tikzpicture}[
%         node distance=3cm,  % Adjusts spacing between nodes
%         ]

%         % Define your flowchart shapes here with styles
%         \tikzstyle{startstop} = [rounded rectangle, draw,  minimum width=2cm, minimum height=1cm, text centered];
%         \tikzstyle{decision} = [diamond, draw,  minimum width=2cm, minimum height=1cm, text centered];
%         \tikzstyle{process} = [rectangle, draw,  minimum width=2cm, minimum height=1cm, text centered];

%         % Add your flowchart elements here
%         \node [startstop] (start) {Input $\Theta^A$, $\Theta^B$};
%         \node [process, below of=start] (process1) {Construct lattice from $\Theta^A$, $\Theta^B$};
%         \node [process, below of=process1] (process2) {Randomly sample lattice node};
%         \node [decision, below of=process2] (decision1) {Is node feasible?};
%         \node [process, right of=process2] (process3) {Discard node (No)};
%         \node [process, below of=decision1] (process4) {Set $\Theta^{A'}$ as the node (Yes)};
%         % \node [startstop, below of=process2] (end_yes) {End (Yes)};
%         \node [startstop, below of=process4] (end_no) {End};

%         % Add arrows to connect the shapes
%         \draw [->] (start) -- (process1);
%         \draw [->] (process1) -- (process2);
%         \draw [->] (process2) -- (decision1);
%         \draw [->] (decision1) -- (process3);
%         \draw [->] (process3) -- (process2);
%         \draw [->] (decision1) -- (process4);
%         \draw [->] (process4) -- (end_no);

%   \end{tikzpicture}
% \end{figure}

Each node in the lattice represents a unique combination of design parameters.
For the purposes of constraint handling, each node in the lattice is evaluated for constraint violations, and the first feasible node to be sampled becomes the offspring.
If no constraints are present, then the first node sampled from the lattice becomes the offspring.

So long as the lattice structure is some function of the distribution of the parents, there are unlimited methods by which a lattice may be structured.
Two approaches for lattice-based crossover will be analyzed in this work.
The first approach is a uniform lattice method, which emulates the functionality of traditional crossover in the N-dimensional space of the linked genes, $\Theta$ \cite{leung2001orthogonal}.
The second approach introduces a Gaussian distributed lattice structure, intended to preserve the population distribution.
Each method is detailed in the following subsections.

% There are two proposed crossover methods (though unlimited possible methods).

\subsubsection{Uniform lattice-based crossover} \label{subsubsec:uniform_lattice_crossover}

The uniform lattice-based approach constructs an even distribution of nodes in the space between the parent chromosomes.
Effectively, the lattice structure is a quantized form of the intermediate recombination \cite{eiben2015introduction}.
% The crossover operation for two continuous decision variables, say $\theta_1$ and $\theta_2$, is a weighted average such that $\theta_{1,offspring} = \theta_1 - r(\theta_1 - \theta_2)$, where $r \in [0, 1]$ is randomly sampled.
% However, the crossover operation for two discrete decision variables merely exchanges the values, with no additional modification (i.e., $r=1$).

Leung and Wang \cite{leung2001orthogonal} details the steps required to construct a uniform lattice.
For brevity, only the salient points are repeated in this work.
The algorithm requires a predetermined number of nodes, $n_p$, which are uniformly distributed in each dimension of the lattice.
The total number of nodes scales exponentially with the cardinality of linked genes, as each gene adds a dimension to the lattice.
For example, a crossover of three linked genes and configured $n_p = 10$ produces a $10 \times 10 \times 10$ lattice, so the lattice contains $n_p^3 = 10^3 = 1000$ nodes.

Following the parlance of \cite{leung2001orthogonal}, the $j^{th}$ level of the $i^{th}$ linked gene is specified by Eq. (\ref{eq:uniform_lattice_nodes}).

\begin{equation} \label{eq:uniform_lattice_nodes}
    \alpha_{i,j} = \Theta_i^A + \left(\frac{j - 1}{n_p - 1}\right)\left(\Theta_i^B - \Theta_i^A\right)
\end{equation}

At the first ($j=1$) and last ($j=n_p$) levels, the node values take on the value of the respective parent gene.
It is required that $n_p \geq 2$.
In the limiting case where $n_p = 2$, the crossover operator reduces to a simple value exchange; this is not recommended, as there are more efficient algorithms that perform value exchange.
Note that the addition and subtraction operators of genes $\Theta_i^A$ and $\Theta_i^B$ may require special attention, such as a modulus function, if the gene domain is bounded.

All the coordinates for a given factor $i$ are then $A_i = [\alpha_{i,1}, \alpha_{i,2}, ..., \alpha_{i,n_p}]$.
In other words, the array $A_i$ contains the discrete values that the $i^{th}$ linked gene may take from the crossover of $\Theta_i^A$ and $\Theta_i^B$.
The procedure for constructing the uniform lattice is summarized in Algorithm \ref{alg:uniform_lattice}.
Algorithm \ref{alg:uniform_lattice} computes the arrays $A_i$ for each linked gene.

\begin{algorithm}
    \hspace*{\algorithmicindent} \textbf{Input}: $\Theta^A$, $\Theta^B$, $n_p$ \\
    \hspace*{\algorithmicindent} \textbf{Output}: $\Theta^{A'}$
    \caption{Uniform Lattice Construction for Parent A}\label{alg:uniform_lattice}\
    \begin{algorithmic}
        \For{$i = 1:N$} \Comment{$N$ is the cardinality of $\Theta^A$ and $\Theta^B$}
            \For{$j = 1:n_p$}
                \State $\alpha_{i,j} \gets \Theta_i^A + \left(\frac{j - 1}{n_p - 1}\right)\left(\Theta_i^B - \Theta_i^A\right)$
                \EndFor
            \State $A_{i} \gets [\alpha_{i,1}, \alpha_{i,2}, ..., \alpha_{i,n_p}]$
        \EndFor
        \State $L \gets \Call{NDimensionalGrid}{A_1, A_2, ..., A_N}$
        \State \Comment{Repeat steps with reversed inputs for $\Theta^{B'}$}
    \end{algorithmic}
\end{algorithm}

% As a pedagogical example, consider the crossover of parents with two lined genes, such that $\Theta^A = [1, 10]$ and $\Theta^B = [2, 12]$.
% For $n_p = 3$,

To form the N-dimensional lattice, each 1-dimensional $A_i$ array is permutated to produce nodes all possible combinations of linked alleles via the $\Call{NDimensionalGrid}$ function.
Example implementations of\linebreak $\Call{NDimensionalGrid}$ are the ``ndgrid'' function in MATLAB \cite{MATLAB} or the ``meshgrid'' function in Python's NumPy package~\cite{harris2020array}.

Figure \ref{fig:grid_lattice_example} shows an example of the uniform lattice structure, with nodes evenly distributed between the parents.
The hatch filled shape signifies an infeasible region in the design space due to a constraint.
The nodes with the infeasible region are shaded orange, which indicates that they are not valid selections for the offspring.
The unshaded nodes are valid choices for the offspring, and all result in alleles that are different than either of the parents.
The parent nodes $\Theta^A$ and $\Theta^B$ are also valid choices for the offspring, which guarantees that the recombination of two feasible parents results in a feasible offspring.
The offspring will be selected randomly from the unshaded nodes or parent nodes, all of which are feasible solutions.

% orange -- {rgb:red,229;green,117;blue,31}
% teal -- {rgb:red,44;green,213;blue,196}
% maroon -- {rgb:red,134;green,31;blue,65}
% pink -- {rgb:red,206;green,0;blue,88}
% hokie stone -- {rgb:red,117;green,120;blue,123}
\usetikzlibrary{arrows.meta}
\begin{figure}
    \centering
    \begin{tikzpicture}

        % Constraint boundary
        \begin{scope}[local bounding box=L]
            % \filldraw[draw={rgb:red,134;green,31;blue,65}, fill={rgb:red,134;green,31;blue,65}] (24pt,-6pt)--(24pt,150pt)--(72pt,150pt)--(72pt,-6pt)--(24pt,-6pt);
            % \filldraw[draw={rgb:red,134;green,31;blue,65}, fill={rgb:red,134;green,31;blue,65}] plot [smooth cycle] coordinates {(24pt,0pt) (24pt,150pt) (70pt,150pt) (130pt,40pt) (72pt,0pt)};
            \filldraw[pattern=north west lines, pattern color={rgb:red,134;green,31;blue,65}] plot [smooth cycle] coordinates {(24pt,0pt) (24pt,150pt) (70pt,150pt) (130pt,40pt) (72pt,0pt)};
            \node[draw={rgb:red,134;green,31;blue,65}, fill={rgb:red,134;green,31;blue,65}, text=white] at (48pt,160pt) {constraint};
        \end{scope}

        % Lattice outer boundary
        \draw (0pt,0pt) -- (288pt,0pt) -- (288pt,144pt) -- (0pt,144pt) -- (0pt,0pt);

        \draw (0pt,24pt) -- (288pt,24pt);
        \draw (0pt,48pt) -- (288pt,48pt);
        \draw (0pt,72pt) -- (288pt,72pt);
        \draw (0pt,96pt) -- (288pt,96pt);
        \draw (0pt,120pt) -- (288pt,120pt);

        % \draw (24pt,0pt) -- (24pt,144pt);
        \draw (48pt,0pt) -- (48pt,144pt);
        % \draw (72pt,0pt) -- (72pt,144pt);
        \draw (96pt,0pt) -- (96pt,144pt);
        % \draw (120pt,0pt) -- (120pt,144pt);
        \draw (144pt,0pt) -- (144pt,144pt);
        % \draw (168pt,0pt) -- (168pt,144pt);
        \draw (192pt,0pt) -- (192pt,144pt);
        % \draw (216pt,0pt) -- (216pt,144pt);
        \draw (240pt,0pt) -- (240pt,144pt);
        % \draw (264pt,0pt) -- (264pt,144pt);

        \node[text=black] at (-14pt,-11pt) {$\Theta^A$};
        \node[text=black] at (24pt,-15pt) {$\alpha_1$};
        \draw[->] (30pt,-15pt) -- (80pt, -15pt);
        \node[text=black] at (-20pt,10pt) {$\alpha_2$};
        \draw[->] (-20pt,20pt) -- (-20pt, 65pt);
        \filldraw[fill={rgb:red,117;green,120;blue,123}, draw=black] (0pt,0pt) circle (5pt);  % Parent 1
        \filldraw[fill=white, draw=black] (0pt,24pt) circle (5pt);
        \filldraw[fill=white, draw=black] (0pt,48pt) circle (5pt);
        \filldraw[fill=white, draw=black] (0pt,72pt) circle (5pt);
        \filldraw[fill=white, draw=black] (0pt,96pt) circle (5pt);
        \filldraw[fill=white, draw=black] (0pt,120pt) circle (5pt);
        \filldraw[fill=white, draw=black] (0pt,144pt) circle (5pt);

        \filldraw[fill={rgb:red,229;green,117;blue,31}, draw=black] (48pt,0pt) circle (5pt);
        \filldraw[fill={rgb:red,229;green,117;blue,31}, draw=black] (48pt,24pt) circle (5pt);
        \filldraw[fill={rgb:red,229;green,117;blue,31}, draw=black] (48pt,48pt) circle (5pt);
        \filldraw[fill={rgb:red,229;green,117;blue,31}, draw=black] (48pt,72pt) circle (5pt);
        \filldraw[fill={rgb:red,229;green,117;blue,31}, draw=black] (48pt,96pt) circle (5pt);
        \filldraw[fill={rgb:red,229;green,117;blue,31}, draw=black] (48pt,120pt) circle (5pt);
        \filldraw[fill={rgb:red,229;green,117;blue,31}, draw=black] (48pt,144pt) circle (5pt);

        \filldraw[fill=white, draw=black] (96pt,0pt) circle (5pt);
        \filldraw[fill={rgb:red,229;green,117;blue,31}, draw=black] (96pt,24pt) circle (5pt);
        \filldraw[fill={rgb:red,229;green,117;blue,31}, draw=black] (96pt,48pt) circle (5pt);
        \filldraw[fill={rgb:red,229;green,117;blue,31}, draw=black] (96pt,72pt) circle (5pt);
        \filldraw[fill={rgb:red,229;green,117;blue,31}, draw=black] (96pt,96pt) circle (5pt);
        \filldraw[fill=white, draw=black] (96pt,120pt) circle (5pt);
        \filldraw[fill=white, draw=black] (96pt,144pt) circle (5pt);

        \filldraw[fill=white, draw=black] (144pt,0pt) circle (5pt);
        \filldraw[fill=white, draw=black] (144pt,24pt) circle (5pt);
        \filldraw[fill=white, draw=black] (144pt,48pt) circle (5pt);
        \filldraw[fill=white, draw=black] (144pt,72pt) circle (5pt);
        \filldraw[fill=white, draw=black] (144pt,96pt) circle (5pt);
        \filldraw[fill=white, draw=black] (144pt,120pt) circle (5pt);
        \filldraw[fill=white, draw=black] (144pt,144pt) circle (5pt);

        \filldraw[fill=white, draw=black] (192pt,0pt) circle (5pt);
        \filldraw[fill=white, draw=black] (192pt,24pt) circle (5pt);
        \filldraw[fill=white, draw=black] (192pt,48pt) circle (5pt);
        \filldraw[fill=white, draw=black] (192pt,72pt) circle (5pt);
        \filldraw[fill=white, draw=black] (192pt,96pt) circle (5pt);
        \filldraw[fill=white, draw=black] (192pt,120pt) circle (5pt);
        \filldraw[fill=white, draw=black] (192pt,144pt) circle (5pt);

        \filldraw[fill=white, draw=black] (240pt,0pt) circle (5pt);
        \filldraw[fill=white, draw=black] (240pt,24pt) circle (5pt);
        \filldraw[fill=white, draw=black] (240pt,48pt) circle (5pt);
        \filldraw[fill=white, draw=black] (240pt,72pt) circle (5pt);
        \filldraw[fill=white, draw=black] (240pt,96pt) circle (5pt);
        \filldraw[fill=white, draw=black] (240pt,120pt) circle (5pt);
        \filldraw[fill=white, draw=black] (240pt,144pt) circle (5pt);

        \filldraw[fill=white, draw=black] (288pt,0pt) circle (5pt);
        \filldraw[fill=white, draw=black] (288pt,24pt) circle (5pt);
        \filldraw[fill=white, draw=black] (288pt,48pt) circle (5pt);
        \filldraw[fill=white, draw=black] (288pt,72pt) circle (5pt);
        \filldraw[fill=white, draw=black] (288pt,96pt) circle (5pt);
        \filldraw[fill=white, draw=black] (288pt,120pt) circle (5pt);
        \filldraw[fill={rgb:red,117;green,120;blue,123}, draw=black] (288pt,144pt) circle (5pt);  % Parent 2
        \node[text=black] at (302pt,157pt) {$\Theta^B$};

    \end{tikzpicture}

    \caption{Uniform lattice, 2-D example. The shaded nodes inside the infeasible region violate the problem constraints and cannot become the offspring.}
    \label{fig:grid_lattice_example}
\end{figure}
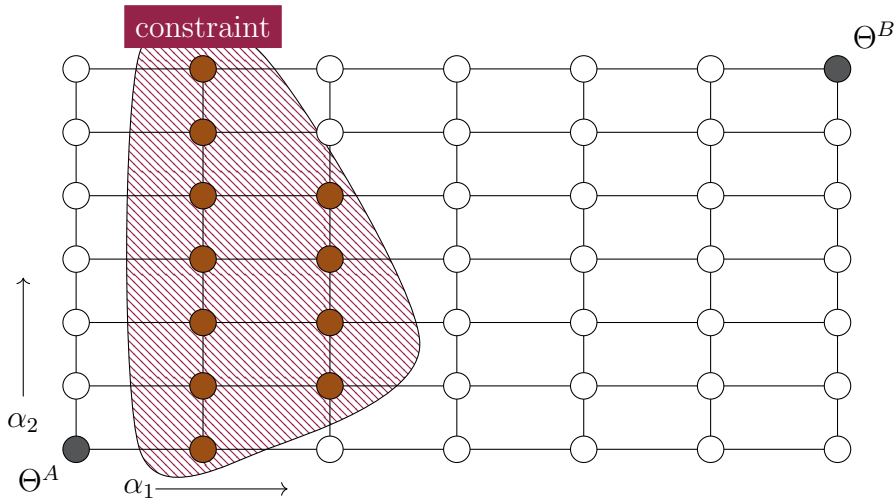

The lattice structure in a two-dimensional quantization is identical to that shown in \cite{leung2001orthogonal}.
Notably, the number of lattice nodes grows exponentially with the number of dimensions.
The primary contribution of \cite{leung2001orthogonal} introduced principles of orthogonal design to cull the points to a representative set, efficiently guide the local search in high dimensions.
In this work, an alternative approach is taken, where the lattice is instead down-selected to only admit feasible solutions.

\subsubsection{Gaussian lattice-based crossover}

The Gaussian lattice-based approach searches in the region surrounding the parent chromosomes, rather than the space enclosed between the parents.
This method of local search may provide an advantage in the presence of constraints, as the region around a parent is likely to satisfy the constraints.
The approach presented here is conceptually similar to Unimodal Normal Distribution crossover (UNDX), and follows the same fundamental guidelines~\cite{ono2003real}.
UNDX implicitly infers population statistics from the separation of the parents in the design space.
The goal of UNDX is to preserve population diversity and leverage epistatic relationships among alleles.
However, the original UNDX algorithm does not consider constraints, and no quantization is involved.

Like UNDX, the Gaussian lattice-based crossover approach samples offspring from a Gaussian distribution, where the scale parameter is related to the distribution of the parents.
The Gaussian lattice is quantized into equal groups of probability, known as quantiles, which are centered on each parent.
% The exact distribution of the lattice nodes about the parent is derived from the inverse CDF of the normal distribution.
The scale parameter of the Gaussian distribution (being the standard deviation, $\sigma$) is derived from the separation of the parent alleles selected for recombination.
The difference between the allele values will be denoted as $\Delta\Theta = [\Delta\theta_0,...,\Delta\theta_N]$.
It is recommended that the standard deviation of the Gaussian distribution be selected such that the difference between parent alleles encompasses most of the probability distribution.
For instance, in two dimensions, the standard deviation $3\sigma_i = \Delta\theta_i$ satisfies this recommendation.

% The 3$\sigma$ bound is the distance between the parents.

The quantile function of the normal distribution, also known as the probit function \cite{bliss1934method}, is provided in Eq. (\ref{eq:probit_function}).
For a normal distribution, the probit function computes the number of standard deviations away from the mean that encompasses a cumulative probability, denoted as $q$.

\begin{equation} \label{eq:probit_function}
    \Call{probit}{q} = \sqrt{2}\mathrm{erf}^{-1}(2q-1), q \in [0,1]
\end{equation}

The probit function in Eq. (\ref{eq:probit_function}) is leveraged in this work to compute the positions of the nodes in the Gaussian lattice, provided the node's quantile.
Note the use of the inverse error function, $\mathrm{erf}^{-1}$, which cannot be evaluated in closed form.

Algorithm \ref{alg:gauss_lattice} outlines the framework required to construct a Gaussian lattice, $L$, for crossover.
The purpose of the algorithm is to construct $L$ such that the density of nodes is Gaussian-distributed about the parent.
Specifically, the lattice nodes will be more densely distributed near the parent, encouraging search there.
The lattice is centered on the parent, so that the central node is guaranteed to be feasible.
The lattice nodes are symmetrically distributed about the center.  % , permitting the offspring to be selected from outside the space enclosed by the parents, distinct from traditional crossover methods.

The Gaussian lattice structure is conceptually similar to the uniform lattice shown in Subsection \ref{subsubsec:uniform_lattice_crossover}, but with a few differences.
Instead of spanning the space between the two parents, the Gaussian lattice is centered on one of the parents and symmetrically spans the space about that parent.
The structural symmetry of the lattice about the parent provides a different utility than the uniform lattice because it enables offspring selection outside of the space enclosed between the parents.

Additionally, the Gaussian lattice structure inherently emphasizes search near the parent;
while the lattice nodes are equal probability for selection, the higher density of nodes near the parent increases the likelihood of selecting an offspring there.
This feature is particularly advantageous for granular search when the parent is near a local minima.
Skew distributions exist that could also be leveraged to generate lattices with more nodes between the parents, but the exploration of such lattice structures was beyond the scope of this work.

\begin{figure}
    \centering
    \begin{tikzpicture}[scale=0.55]

        % Inner circle tiles
        \filldraw[fill=white, fill opacity=0.0, draw=black] (0pt, 0pt) to (19.9pt, 48.0pt) to [out=157.5, in=22.5] (-19.9pt, 48.0pt) to (0pt, 0pt);
        \filldraw[fill=white, fill opacity=0.0, draw=black] (0pt, 0pt) to (-19.9pt, 48.0pt) to [out=202.5, in=67.5] (-48.0pt, 19.9pt) to (0pt, 0pt);
        \filldraw[fill=white, fill opacity=0.0, draw=black] (0pt, 0pt) to (-48.0pt, 19.9pt) to [out=247.5, in=112.5] (-48.0pt, -19.9pt) to (0pt, 0pt);
        \filldraw[fill=white, fill opacity=0.0, draw=black] (0pt, 0pt) to (-48.0pt, -19.9pt) to [out=292.5, in=157.5] (-19.9pt, -48.0pt) to (0pt, 0pt);
        \filldraw[fill=white, fill opacity=0.0, draw=black] (0pt, 0pt) to (-19.9pt, -48.0pt) to [out=337.5, in=202.5] (19.9pt, -48.0pt) to (0pt, 0pt);
        \filldraw[fill=white, fill opacity=0.0, draw=black] (0pt, 0pt) to (19.9pt, -48.0pt) to [out=22.5, in=247.5] (48.0pt, -19.9pt) to (0pt, 0pt);
        \filldraw[fill=white, fill opacity=0.0, draw=black] (0pt, 0pt) to (48.0pt, -19.9pt) to [out=67.5, in=292.5] (48.0pt, 19.9pt) to (0pt, 0pt);
        \filldraw[fill=white, fill opacity=0.0, draw=black] (0pt, 0pt) to (48.0pt, 19.9pt) to [out=112.5, in=337.5] (19.9pt, 48.0pt) to (0pt, 0pt);

        % Middle circle tiles
        \filldraw[fill=white, fill opacity=0.0, draw=black] (19.9pt, 48.0pt) to [out=157.5, in=22.5] (-19.9pt, 48.0pt) to (-42.1pt, 101.6pt) to [out=22.5, in=157.5] (42.1pt, 101.6pt) to (19.9pt, 48.0pt);
        \filldraw[fill=white, fill opacity=0.0, draw=black] (-19.9pt, 48.0pt) to [out=202.5, in=67.5] (-48.0pt, 19.9pt) to (-101.6pt, 42.1pt) to [out=67.5, in=202.5] (-42.1pt, 101.6pt) to (-19.9pt, 48.0pt);
        \filldraw[fill=white, fill opacity=0.0, draw=black] (-48.0pt, 19.9pt) to [out=247.5, in=112.5] (-48.0pt, -19.9pt) to (-101.6pt, -42.1pt) to [out=112.5, in=247.5] (-101.6pt, 42.1pt) to (-48.0pt, 19.9pt);
        \filldraw[fill=white, fill opacity=0.0, draw=black] (-48.0pt, -19.9pt) to [out=292.5, in=157.5] (-19.9pt, -48.0pt) to (-42.1pt, -101.6pt) to [out=157.5, in=292.5] (-101.6pt, -42.1pt) to (-48.0pt, -19.9pt);
        \filldraw[fill=white, fill opacity=0.0, draw=black] (-19.9pt, -48.0pt) to [out=337.5, in=202.5] (19.9pt, -48.0pt) to (42.1pt, -101.6pt) to [out=202.5, in=337.5] (-42.1pt, -101.6pt) to (-19.9pt, -48.0pt);
        \filldraw[fill=white, fill opacity=0.0, draw=black] (19.9pt, -48.0pt) to [out=22.5, in=247.5] (48.0pt, -19.9pt) to (101.6pt, -42.1pt) to [out=247.5, in=22.5] (42.1pt, -101.6pt) to (19.9pt, -48.0pt);
        \filldraw[fill=white, fill opacity=0.0, draw=black] (48.0pt, -19.9pt) to [out=67.5, in=292.5] (48.0pt, 19.9pt) to (101.6pt, 42.1pt) to [out=292.5, in=67.5] (101.6pt, -42.1pt) to (48.0pt, -19.9pt);
        \filldraw[fill=white, fill opacity=0.0, draw=black] (48.0pt, 19.9pt) to [out=112.5, in=337.5] (19.9pt, 48.0pt) to (42.1pt, 101.6pt) to [out=337.5, in=112.5] (101.6pt, 42.1pt) to (48.0pt, 19.9pt);

        % Large circle tiles
        \filldraw[fill=white, fill opacity=0.0, draw=black] (42.1pt, 101.6pt) to [out=157.5, in=22.5] (-42.1pt, 101.6pt) to (-76.5pt, 184.8pt) to [out=22.5, in=157.5] (76.5pt, 184.8pt) to (42.1pt, 101.6pt);
        \filldraw[fill=white, fill opacity=0.0, draw=black] (-42.1pt, 101.6pt) to [out=202.5, in=67.5] (-101.6pt, 42.1pt) to (-184.8pt, 76.5pt) to [out=67.5, in=202.5] (-76.5pt, 184.8pt) to (-42.1pt, 101.6pt);
        \filldraw[fill=white, fill opacity=0.0, draw=black] (-101.6pt, 42.1pt) to [out=247.5, in=112.5] (-101.6pt, -42.1pt) to (-184.8pt, -76.5pt) to [out=112.5, in=247.5] (-184.8pt, 76.5pt) to (-101.6pt, 42.1pt);
        \filldraw[fill=white, fill opacity=0.0, draw=black] (-101.6pt, -42.1pt) to [out=292.5, in=157.5] (-42.1pt, -101.6pt) to (-76.5pt, -184.8pt) to [out=157.5, in=292.5] (-184.8pt, -76.5pt) to (-101.6pt, -42.1pt);
        \filldraw[fill=white, fill opacity=0.0, draw=black] (-42.1pt, -101.6pt) to [out=337.5, in=202.5] (42.1pt, -101.6pt) to (76.5pt, -184.8pt) to [out=202.5, in=337.5] (-76.5pt, -184.8pt) to (-42.1pt, -101.6pt);
        \filldraw[fill=white, fill opacity=0.0, draw=black] (42.1pt, -101.6pt) to [out=22.5, in=247.5] (101.6pt, -42.1pt) to (184.8pt, -76.5pt) to [out=247.5, in=22.5] (76.5pt, -184.8pt) to (42.1pt, -101.6pt);
        \filldraw[fill=white, fill opacity=0.0, draw=black] (101.6pt, -42.1pt) to [out=67.5, in=292.5] (101.6pt, 42.1pt) to (184.8pt, 76.5pt) to [out=292.5, in=67.5] (184.8pt, -76.5pt) to (101.6pt, -42.1pt);
        \filldraw[fill={rgb:red,80;green,133;blue,144}, fill opacity=1.0, draw=black] (101.6pt, 42.1pt) to [out=112.5, in=337.5] (42.1pt, 101.6pt) to (76.5pt, 184.8pt) to [out=337.5, in=112.5] (184.8pt, 76.5pt) to (101.6pt, 42.1pt);

        \node[above] at (111.6pt, 50.1pt) {4\%};
        % Draw circles at each quantile

        %% Draw feasible circles

        % Inner circle
        \filldraw[fill=white, draw=black] (0,24pt) circle (5pt);
        \filldraw[fill=white, draw=black] (0,-24pt) circle (5pt);
        \filldraw[fill={rgb:red,229;green,117;blue,31}, draw=black] (-24pt,0) circle (5pt);
        \filldraw[fill=white, draw=black] (24pt,0) circle (5pt);
        \filldraw[fill=white, draw=black] (16.97pt,16.97pt) circle (5pt);
        \filldraw[fill=white, draw=black] (16.97pt,-16.97pt) circle (5pt);
        \filldraw[fill=white, draw=black] (-16.97pt,16.97pt) circle (5pt);
        \filldraw[fill=white, draw=black] (-16.97pt,-16.97pt) circle (5pt);

        % Middle circle
        \filldraw[fill=white, draw=black] (0,76pt) circle (5pt);
        \filldraw[fill=white, draw=black] (0,-76pt) circle (5pt);
        \filldraw[fill={rgb:red,229;green,117;blue,31}, draw=black] (-76pt,0) circle (5pt);
        \filldraw[fill=white, draw=black] (76pt,0) circle (5pt);
        \filldraw[fill=white, draw=black] (53.74pt,53.74pt) circle (5pt);
        \filldraw[fill=white, draw=black] (53.74pt,-53.74pt) circle (5pt);
        \filldraw[fill={rgb:red,229;green,117;blue,31}, draw=black] (-53.74pt,53.74pt) circle (5pt);
        \filldraw[fill=white, draw=black] (-53.74pt,-53.74pt) circle (5pt);

        % Large circle
        \filldraw[fill=white, draw=black] (0,144pt) circle (5pt);
        \filldraw[fill=white, draw=black] (0,-144pt) circle (5pt);
        \filldraw[fill=white, draw=black] (-144pt,0) circle (5pt);
        \filldraw[fill=white, draw=black] (144pt,0) circle (5pt);
        \filldraw[fill=white, draw=black] (101.82pt,101.82pt) circle (5pt);
        \filldraw[fill=white, draw=black] (101.82pt,-101.82pt) circle (5pt);
        \filldraw[fill={rgb:red,229;green,117;blue,31}, draw=black] (-101.82pt,101.82pt) circle (5pt);
        \filldraw[fill=white, draw=black] (-101.82pt,-101.82pt) circle (5pt);

        \begin{scope}[local bounding box=L]
            \filldraw[pattern=north west lines, pattern color={rgb:red,134;green,31;blue,65}] plot [smooth cycle] coordinates {(-24pt,-30pt) (-24pt,150pt) (-74pt,150pt) (-140pt,20pt) (-72pt,-30pt)};
            \node[draw={rgb:red,134;green,31;blue,65}, fill={rgb:red,134;green,31;blue,65}, text=white] at (-67pt,160pt) {constraint};
        \end{scope}

        \filldraw[fill={rgb:red,117;green,120;blue,123}, draw=black] (0,0) circle (5pt);
        \node[text=black, fill=white, draw=black] at (10pt,-20pt) {\scriptsize $\Theta^A$};
        \filldraw[fill={rgb:red,117;green,120;blue,123}, draw=black] (198.5092pt,24.3739pt) circle (5pt);
        \node[text=black] at (218pt,25pt) {\scriptsize $\Theta^B$};

        \node[text=black] at (0pt,-220pt) {$\gamma_1$};
        \draw[->] (10pt,-220pt) -- (80pt, -220pt);
        \node[text=black] at (-220pt,0pt) {$\gamma_2$};
        \draw[->] (-220pt,10pt) -- (-220pt, 80pt);

    \end{tikzpicture}
    \caption{Gaussian lattice 2-D example. The shaded nodes inside the infeasible region violate the problem constraints and cannot become the offspring. }
    \label{fig:gauss_lattice_example}
\end{figure}
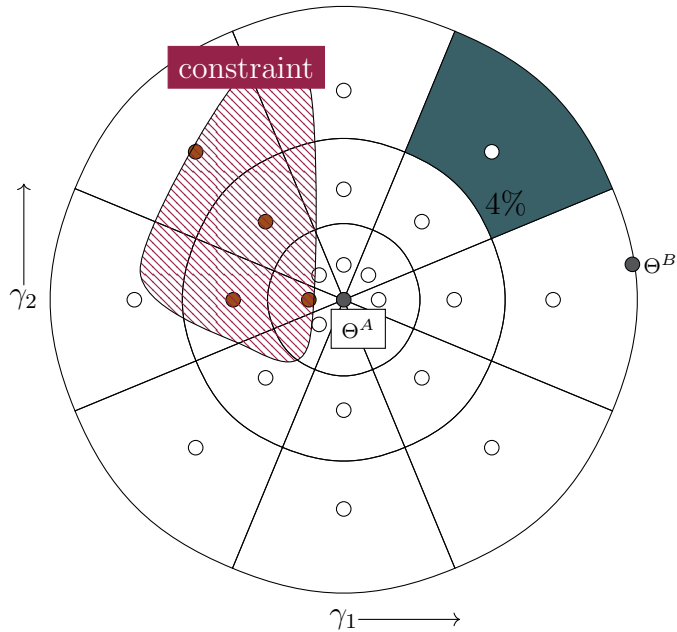

The premise of the Gaussian lattice is that a specified number of points, $n_p$, are placed uniformly at each quantile of the Gaussian distribution centered on the parent.
% The standard deviation of the Gaussian distribution is determined from the space bounded by the genes of Parent A and Parent B.
Because the lattice may exist in N dimensions, the general procedure for lattice construction is to evenly distribute nodes at each quantile along the surface of a hypersphere.
Note that the hypersphere radius is determined by the quantile of the distribution.
For two linked genes, the lattice structure is a sequence of 1-spheres, as depicted in Fig. \ref{fig:gauss_lattice_example}.
For three linked genes, the lattice structure is a sequence of 2-spheres, and the Fibonacci algorithm can approximate an even distribution of nodes for the spheres.
In higher dimensions, no straightforward solution exists to evenly distribute nodes on a hypersphere.
However, approximations of the higher dimensional Fibonacci algorithm exist, and are suitable for lattice-based crossover~\cite{lovisolo2001uniform}.  % https://math.stackexchange.com/questions/3291489/can-the-fibonacci-lattice-be-extended-to-dimensions-higher-than-3.
A simpler alternative may be to randomly sample the desired number of lattice points from a multivariate Gaussian distribution \cite{muller1959note, marsaglia1972choosing}, but this alternative approach is not considered in this work.

To construct the lattice, the positions of the $n_p$ lattice nodes are determined for a unit hypersphere.
Let $S$ denote a set of coordinates that sit on a unit hypersphere.
$S$ is composed of several points of unit radius, $s$, as defined in Eq. (\ref{eq:unit_hypersphere}).

\begin{equation} \label{eq:unit_hypersphere}
    S = [s_0,...,s_{n_p}], \forall s \in \mathbb{R}^N : \lVert s \rVert = 1
\end{equation}

Once each $s \in S$ is known, the coordinates can be scaled to a hypersphere of arbitrary radius.
In the context of the Gaussian lattice, the radius of each hypersphere that comprises the lattice is determined from the quantile.

% % The points along the surface of a hypersphere at a given quantile will be of radius $\gamma$.
% Each dimension of the coordinate $s$ corresponds a different allele $\theta$ in the linked set of alleles $\Theta$.
% For instance, in a linked set of two alleles (i.e., $\Theta = [\theta_1, \theta_2]$), $s$ will be described by two coordinates,
% https://www.sagepub.com/sites/default/files/upm-binaries/42772_1_Introduction_to_Statistics.pdf
The general procedure to form the Gaussian lattice is to map the nodes along the unit hypersphere to allele values at a given quantile.
Recall in this work, the difference between the Parent A and Parent B alleles, $\Delta\Theta$, is treated as the $3\sigma$ bound for the Gaussian lattice.
% The possible allele values in a Gaussian lattice are a function of the distribution
Let $\bm{\Gamma}$ be an array of values with the same cardinality as $\Delta\Theta$ which scales the nodes of the unit hypersphere $S$ to the desired hypersphere at quantile $q$.
For a given quantile $q$, the calculation for $\bm{\Gamma}$ is provided in Eq. (\ref{eq:gamma}).

\begin{equation} \label{eq:gamma}
    \bm{\Gamma} = \frac{\Delta\Theta}{3} \Call{probit}{q}
\end{equation}

The general procedure is to compute $\bm{\Gamma}$ from the quantile $q$, and element-wise scale each $s \in S$ by $\bm{\Gamma}$ to form the nodes at $q$.
Intuitively, $\bm{\Gamma}$ represents the radius of a hypersphere at $q$, but $\bm{\Gamma}$ is only scalar if the difference between two parent chromosomes $\Delta\Theta$ is also scalar.
This special case where $\Delta\Theta$ is scalar-valued results when the linked genes have defined equidistant scales, such as genes that represent coordinates within a coordinate system.
In such cases, $\bm{\Gamma}$ also takes on a scalar value, and multiples all elements of each $s$ equally for a given $q$.
Section \ref{subsec:geographic_search} provides an example of this special case.

If design space of $\theta_i$ is Euclidean, then a lattice point is simply $\theta_i + \bm{\Gamma}_i s_i$.
For the non-Euclidean case, other parameters may be needed to determine the coordinates of a lattice point.
For this reason, the complete set of arguments in the signature of $\Call{SamplePoint}$ remains ambiguous in Algorithm \ref{alg:gauss_lattice}.

\begin{algorithm}
    \hspace*{\algorithmicindent} \textbf{Input}: $\Theta_A$, $\Theta_B$, $n_p$, $n_q$  \\
    \hspace*{\algorithmicindent} \textbf{Output}: $L$
    \caption{Gauss Lattice Construction for Parent A}\label{alg:gauss_lattice}\
    \begin{algorithmic}
        \State $\Delta\Theta \gets | \Theta_A - \Theta_B  |$ \Comment{$\Theta$ has shape $1 \times N$}
        \State $S \gets \Call{ConstructUnitHypersphere}{n_p}$  \Comment{$S$ has shape $n_p \times N$}
        \State $M \gets n_p \times n_q$  \Comment{$M$ is the total number of lattice nodes}
        \State $L \gets$ Allocate with shape $M \times N$
        \For{$i = 1:n_q$}  \Comment{Iterate through quantiles}
            \State $\gamma \gets \frac{\Delta\Theta}{3} \Call{probit}{\frac{i}{n_q + 1}}$
            \For{$j = 1:n_p$}  \Comment{Iterate through nodes at quantile}
                \State $s \gets S_{j}$
                \State $k \gets \left(i-1\right)n_q + j$
                \State $L_k \gets \Call{SamplePoint}{\Theta_A, \gamma, s, ...}$
            \EndFor
        \EndFor
    \end{algorithmic}
\end{algorithm}

Figure \ref{fig:gauss_lattice_example} shows an example two-dimensional Gaussian lattice structure, with $n_p = 8$ nodes symmetrically distributed at $n_q = 3$ quantiles.
As with the uniform lattice structure in Fig. \ref{fig:grid_lattice_example} the hatch filled shape signifies an infeasible region, and the orange shaded nodes are invalid choices for the offspring.
Distinct from the uniform lattice, only one parent node (in this illustration $\Theta^A$) is included in the lattice.
The presence of at least one parent in the lattice still guarantees a feasible offspring.
The offspring will be selected randomly from the unshaded nodes or parent node $\Theta^A$, all of which are feasible solutions.
$\Theta^B$ is shown in a the diagram for contextual understanding, but $\Theta^B$ is not considered as a node in the lattice.

In the illustration, each tile that encloses a node occupies an equal amount of probability.
For the 24 tiles shown, each tile encloses about 4\% of the probability distribution.

\subsection{Mutation}  \label{subsec:mutation}

Mutation is the stochastic modification to a chromosome, which encourages global search.
The mutation operator randomly selected genes within the chromosome and overwrites the alleles with different values.
Traditionally, when a gene mutates, the modified allele can take on a value anywhere within the gene's domain.
If search is to be restricted only to feasible regions, then the mutation operator must be adapted accordingly.

In a context where genes are linked, random modifications to individual genes may result in infeasible solutions.
Therefore if a gene is selected for mutation and it is linked with other genes, all linked genes must be modified to ensure the candidate remains in and fully explores the feasible space.
Two mutation methods are suggested here for ensured solution feasibility: (1) realtime resampling and (2) advance sampling.

In the case of realtime resampling, all linked genes are recomputed at runtime when a linked gene is mutated.
The linked set of genes must be repeatedly recomputed until the candidate is feasible, which ensures that no constraints are violated by the mutated chromosome.
A drawback of realtime resampling is that it may be slow if there are many genes linked together or if most of the design space is infeasible.
Additionally, realtime resampling may rarely produce candidates in small feasible regions surrounded by large infeasible regions (e.g., as remote islands in a geographic search problem) due to the low probability of sampling there.

% An alternative approach is used in this work.
In the case of advance sampling, a large but finite set of feasible combinations of linked genes is computed prior to the start of the GA.
When a linked gene is mutated, the new set of linked genes is selected from the precomputed set.
The precomputed set may be computed manually, or by random sampling, or some combination of the two.
While advance sampling precludes mutation from truly performing global search, a dense set of precomputed feasible allele combinations should be sufficient.
Additionally, the advance sampling approach offers a unique advantage.
Known solution candidates of interest may be ``preempted'' by manually inserting them into the precomputed set, increasing the likelihood of search there.
Advance sampling is used in this work.

\section{Results}\label{sec:results}

\subsection{Geographic search performance characterization} \label{subsec:geographic_search}

Revisiting the problem of geographic search, the parameters that determine geographic location are the combination of latitude, $\phi$, and longitude, $\lambda$.
The results from Section \ref{sec:motivation} will be compared with two lattice-based approaches:

\begin{enumerate}
    % \item Repair
    \item Uniform lattice
    \item Gaussian lattice
\end{enumerate}

\subsubsection{Lattice-based crossover setup}

\paragraph{Uniform lattice}

In the lattice approach, allow set of linked genes to be $\Theta = [\phi, \lambda]$.
Following Eq. (\ref{eq:uniform_lattice_nodes}), the difference between the longitude genes of two parents, $\lambda_A$ and $\lambda_B$, is wrapped to the domain $(-180\degree, 180\degree]$.
When computing the lattice node coordinates, the latitude values were wrapped to the domain $[-90\degree, 90\degree]$.

\paragraph{Gaussian lattice}

Latitude and longitude are both coordinates a system with equidistant scales, so a scalar $\gamma = \Delta\Theta$ is preferred to measure the arclength between geographic locations.
While the arclength from the center of the lattice to a node is determined from the arclength sizing parameter, the node must be mapped back to latitude-longitude by inverting the haversine formula.
For each node in the lattice, the change in latitude from the parent must be determined first.

Algorithm \ref{alg:get_latlon_point} shows the implementation of the $\Call{SamplePoint}$ function for the special case of latitude-longitude genes.
The first step is to compute the change in latitude, $\phi_2 - \phi_1$.
Once the latitude change is known, the haversine formula may be inverted to compute the change in longitude, $\lambda_2 - \lambda_1$.
Notably, the inverse haversine returns only the magnitude of the difference in longitude, $\lvert \lambda_2 - \lambda_1 \rvert$.
The sign of the difference is determined by the sign on the corresponding coordinate of the unit hypersphere, $s$.

\begin{algorithm}
    \caption{{$\protect\Call{SamplePoint}$} for latitude-longitude genes} \label{alg:get_latlon_point}
    \begin{algorithmic}
        \Procedure{SamplePoint}{$\gamma, s, \phi_{parent}, \lambda_{parent}$}
            \State $\phi_{offspring} \gets \phi_{parent} + {\gamma}s_1$
            \State $\Delta \lambda \gets \cos^{-1}\left(1 + \frac{\cos(\gamma) - \cos(\phi_{offspring} - \phi_{parent})}{\cos(\phi_{offspring})\cos(\phi_{parent})}\right)$
            \State $\lambda_{offspring} \gets \lambda_{parent} + \Call{sign}{s_2}\Delta \lambda$
        \EndProcedure
    \end{algorithmic}
\end{algorithm}

\subsubsection{Monte Carlo analysis of performance}

A Monte Carlo study was performed, where a feasible location was randomly selected on land as the optimal, and the genetic algorithm was tasked to find the optimal location.
The fitness function was simply the distance from the optimal point, measured by the haversine formula in Eq. (\ref{eq:haversine_formula}).
One thousand ``optimal'' points were sampled.
For each point, the genetic algorithm ran three separate times.

\begin{figure}[!ht]
    \centering
    \includegraphics[width=0.7\textwidth]{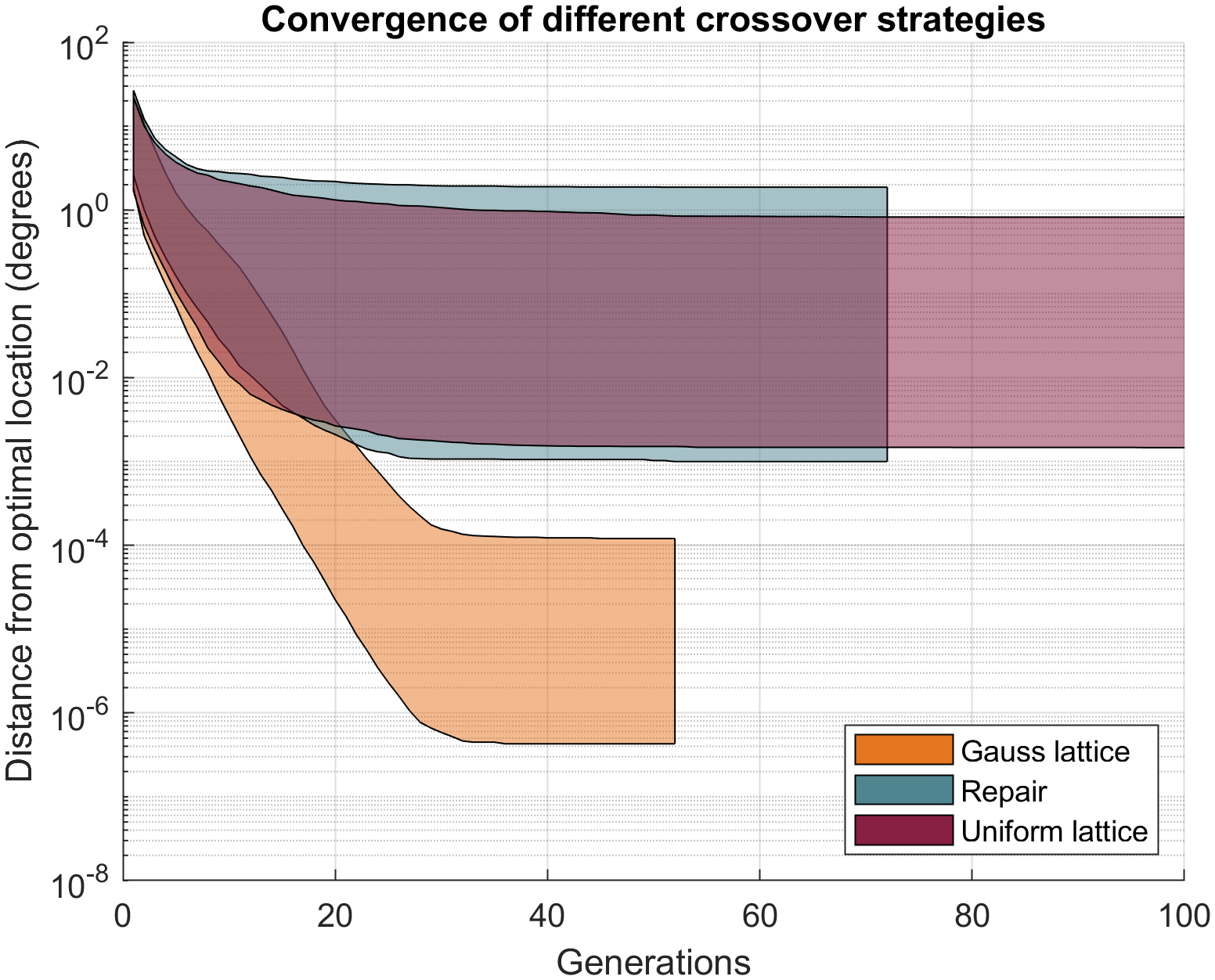}
    \caption{Solution quality of crossover approaches. The upper bound of the shaded region is the ninety-fifth percentile of solutions, and the lower bound is the fifth percentile of solutions.}
    \label{fig:convergence}
\end{figure}

Figure \ref{fig:convergence} shows the results of the Monte Carlo study.
The upper bound of the shaded region demarcates the ninety-fifth percentile of the fitness values for the corresponding technique.
The lower bound of the shaded region demarcates the fifth percentile.
The extent of each shaded region along the horizontal axis is set by the maximum number of generations sustained by the respective crossover operator.
The number of generations shown is truncated at 100.

Notably, the ninety-fifth percentile of the fitness using Gaussian lattice operator is an order of magnitude lower than the fifth percentile of the traditional repair method.
This result is intuitive -- the Gaussian lattice approach is well posed for granular search around minima, since there is a greater likelihood of selecting offspring near the parents.
The sensitivity to epistatic relationships among genes also produces tighter solution precision and faster performance in this application.

\begin{figure}[!ht]
    \centering
    \includegraphics[width=0.7\textwidth]{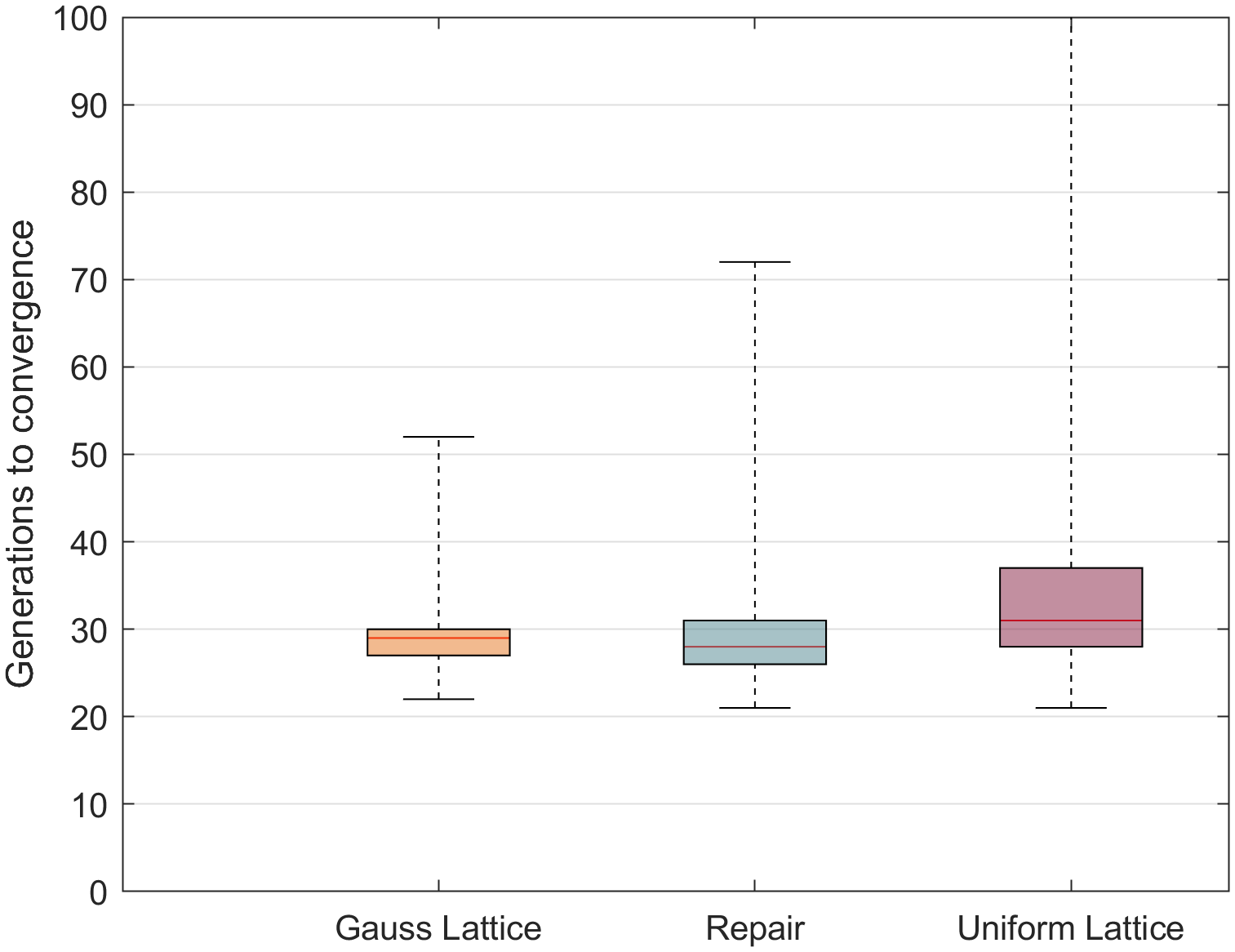}
    \caption{Statistics of generations to convergence of different crossover approaches.}
    \label{fig:gens_convergence}
\end{figure}

Figure \ref{fig:gens_convergence} further illustrates the convergence statistics for each approach.
Notably, the median number of generations to converge is nearly equivalent, but the variance is lowest for the Gaussian lattice-based crossover approach.
While some runs of repair and the uniform lattice-based crossover converged faster than the Gaussian lattice, Fig. \ref{fig:convergence} indicates convergence was premature in these cases.

In this case, the uniform lattice-based crossover does not have a significant performance improvement over repair.
The negligible performance difference is likely because the uniform lattice is functionally similar to traditional crossover methods, where the choices for recombination are restricted to the space enclosed between the parent alleles.
The uniform lattice is not as sensitive to epistatic relationships between genes.
However, if a more traditional approach is preferred, the generality of the lattice method enables the uniform lattice-based crossover to be used in applications where repair is impracticable.

\begin{figure}[ht!]
    \centering
    \includegraphics[width=0.7\textwidth]{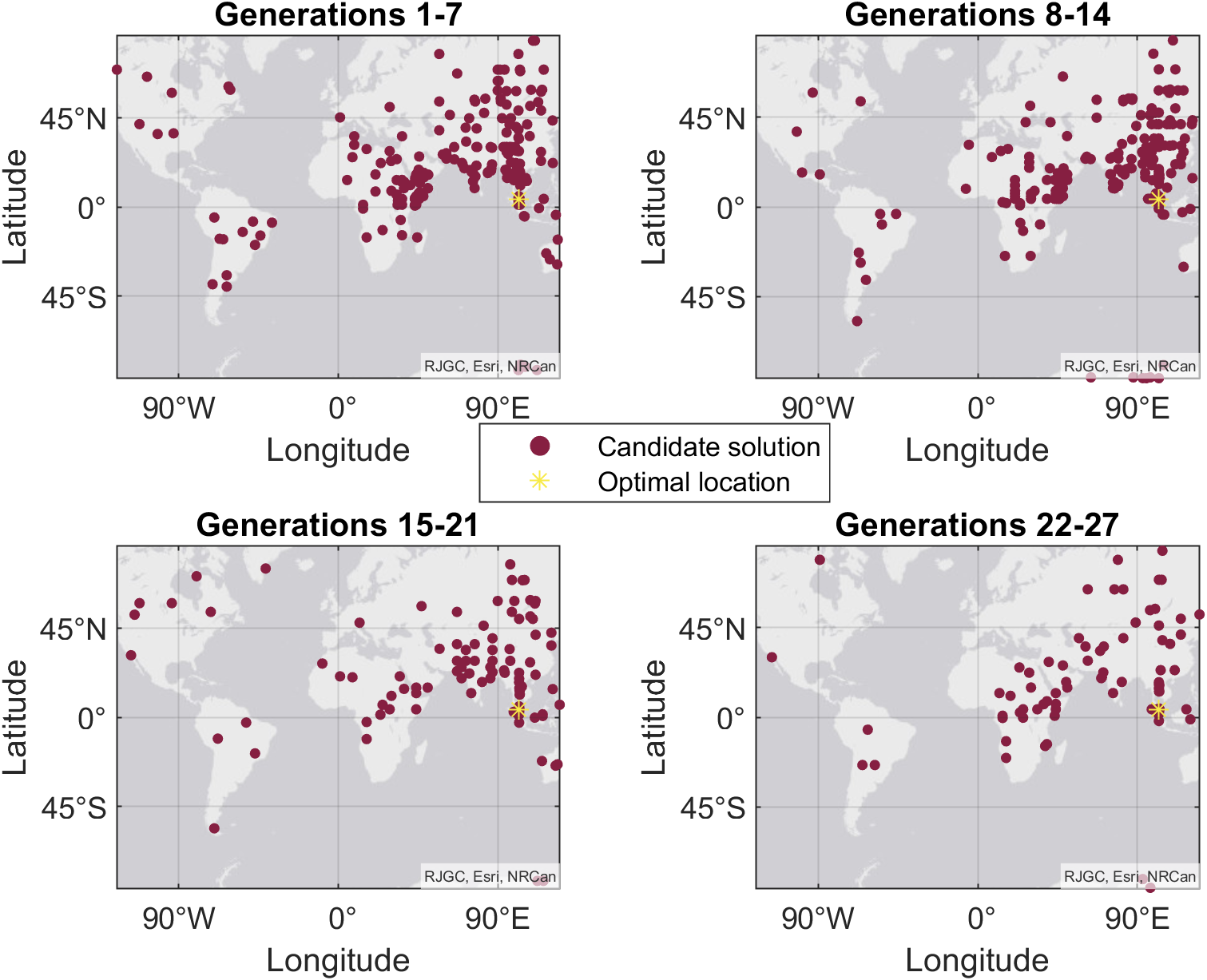}
    \caption{Uniform lattice population evolution. Compared with the solution repair approach, diversity is maintained as the population evolves.}
    \label{fig:uniform_frames}
\end{figure}

While the uniform lattice did not outperform the repair method, its exploration of the design space was notably better when compared to the traditional approach.
An example of the solution diversity of the uniform lattice method over time is shown in Fig. \ref{fig:uniform_frames}.
Note that even as the solutions cluster around the global optimum, other solutions in other parts of the design space, unlike that shown earlier in Fig. \ref{fig:repair_frames}.

\begin{figure}[ht!]
    \centering
    \includegraphics[width=0.7\textwidth]{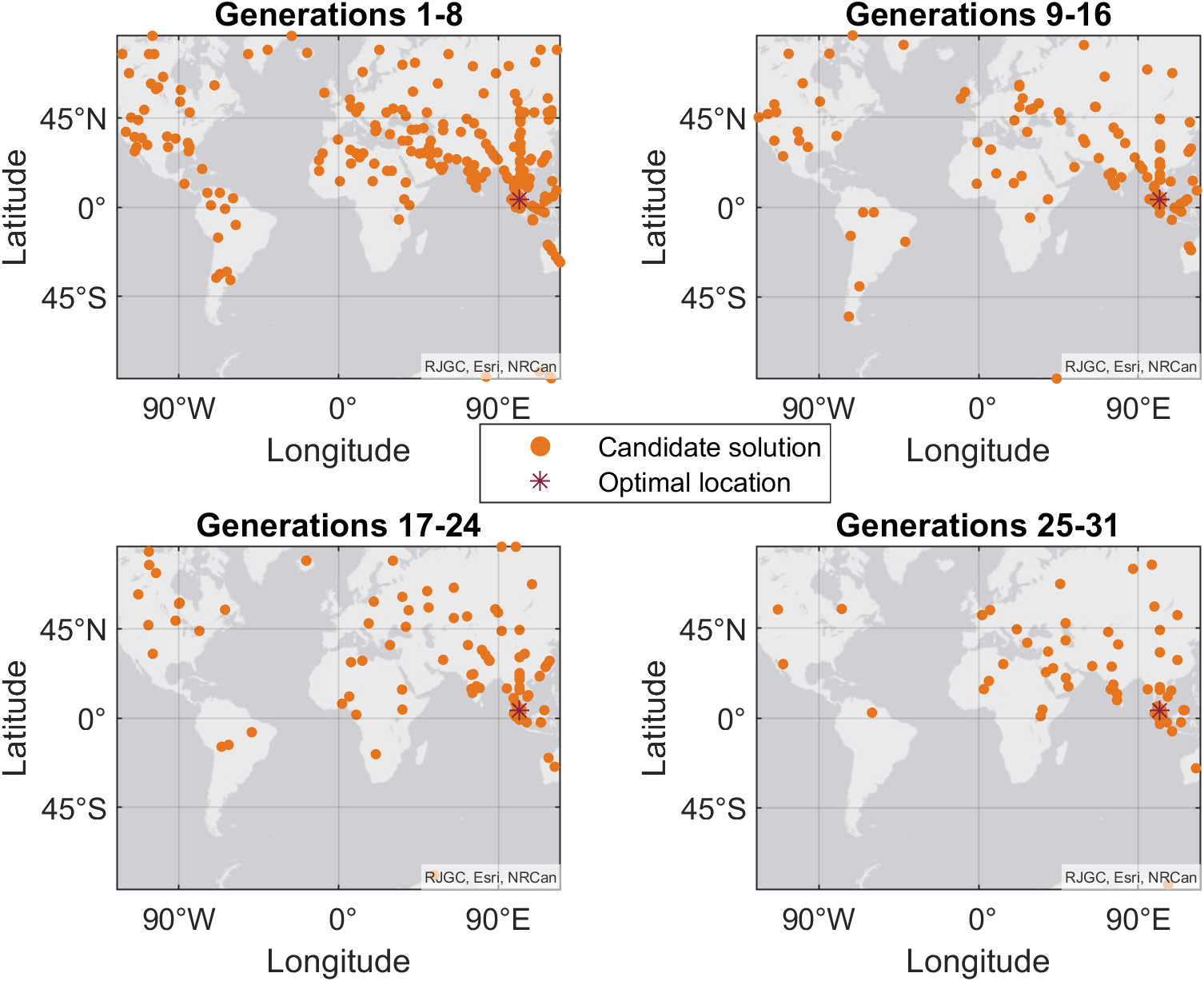}
    \caption{Gaussian lattice population evolution. Compared with the solution repair approach, diversity is maintained as the population evolves.}
    \label{fig:gauss_frames}
\end{figure}

Figure \ref{fig:gauss_frames} illustrates an example of a population's evolution using the Gauss lattice method.
Like the uniform lattice method, the solutions remain spread throughout the design space as the population evolves.

The spokes of the Gauss lattice structure result in a visible spoke pattern around the optimal location in Fig. \ref{fig:gauss_frames}.
The exact number of spokes is determined by the configuration parameter for the number of nodes at each quantile in the lattice structure, $n_q$.
A higher setting for $n_q$ would diminish the spokes but result in more nodes in the lattice, which may impact performance.
Also note that the swarm of solution candidates about the global minimum causes the spoke pattern to appear so sharply.
A space with many Pareto efficient solutions would not feature the same swarming behavior.

The lattice-based crossover method will now be applied in two more practical examples to demonstrate its use.
The superior performance of the Gaussian lattice-based crossover shows considerable promise, and will be the primary focus of the following sample problems.

\subsection{Applied geographic search sample problem}

A modern and relevant example of a geographic search problem can be found in astrophotography.
Astrophotographers must expose optical equipment for many nights to produce a resolved image of a distant celestial body.
Astrophotographers prefer regions with low urban light pollution, to mitigate background noise that could smother dim objects.
A secondary obstacle to astrophotography is cloud cover.
Clouds are opaque to visible light, and preclude an otherwise suitable night of collection.
An optimal location problem for astrophotography will be posed with these objectives.

\subsubsection{Objectives}

\paragraph{Artificial sky brightness}

Light pollution is the product of artificial light sources, which is particularly severe near densely urban areas.
The atmosphere reflects some of the light, which propagates back to the ground.
The reflected light produces noise on the focal plane of a skyward telescope, which may overpower the light originating from space objects.
% This is an issue for space domain awareness because small debris or objects in distant orbits appear dim to terrestrial observers.
Resultant noise from light pollution inhibits observation of dim objects nearer to Earth or celestial phenomena in distant space.

% Geographically distributed zenith sky brightness data is publicly available from \url{https://www.lightpollutionmap.info/}.

Geographically distributed zenith sky brightness data is publicly available courtesy of \cite{falchi2016supplement,falchi2016new}.
A visualization of the data is accessible at \cite{lightpollutionmap}.
The artificial sky brightness data is provided as luminance values, $L_{artificial}$, expressed in millicandelas per square meter over a 30 arcsecond grid.
\cite{kyba2017converting} notes that the data can be transformed into radiometric data with Eq. (\ref{eq:luminance2brightness}).
Note that a value for natural sky brightness must be added to the values from the Atlas -- \cite{kyba2017converting} suggests $0.236 mcd/m^2$.

\begin{equation}\label{eq:luminance2brightness}
    M_{artificial} = -2.5\log{\frac{L_{artificial}}{10.8e7}}
\end{equation}

The luminance data may be interpolated for accurate sky brightnesses across the globe, using MATLAB $\Call{interp2}$ function \cite{MATLAB}.
Interpolation yields luminance as a function of latitude and longitude, $L_{artificial}(\phi, \lambda)$.
Equation (\ref{eq:luminance2brightness}) may then be used to transform the result to radiometric brightness as a function of latitude and longitude, $M_{artificial}(\phi, \lambda)$.

\paragraph{Cloud cover}

Regional weather affects telescope operation on a nightly basis.
The most direct effect of weather on telescope operation is cloud cover.
Sufficiently thick clouds are opaque to visible light, and frequent cloudy nights will cripple a telescope's ability to view the night sky.
While daily weather is random, cloud data over long timescales can inform which regions provide ideal viewing conditions on average.

For decades, the International Satellite Cloud Climatology Project (ISCCP) collected satellite imagery of Earth's cloud cover and recorded the data \cite{noaa2016data}.
One dataset provided by the ISCCP is the H-series Gridded Monthly data, which includes percentage cloud cover data gridded over latitude and longitude.
The data is publicly available, and may be used to recover average cloud cover for a specific location.
For simplicity, the data will be averaged over the course of the most recent year available at the time of writing.
The data is provided as a grid, and may be interpolated using MATLAB's $\Call{interp2}$ function \cite{MATLAB}.

For a given latitude $\phi$ and longitude $\lambda$, the yearly average cloud cover at the location ($\phi$, $\lambda$) is $\overline{O}(\phi, \lambda)$, provided in Eq. (\ref{eq:cloud_objective}).

\begin{equation} \label{eq:cloud_objective}
    \overline{O}(\phi, \lambda) = \frac{1}{12} \sum_{i=1}^{12} O_i(\phi, \lambda)
\end{equation}

where the set $O(\phi, \lambda)$ is the monthly cloud data provided by ISCCP:

\begin{equation*}
    O(\phi, \lambda) = \left[o_{january}(\phi, \lambda), o_{february}(\phi, \lambda), ..., o_{november}(\phi, \lambda), o_{december}(\phi, \lambda)\right]
\end{equation*}

\paragraph{Proximity to owners/operators}

Astrophotography is popular among academics and hobbyists alike.
While hobbyists may be willing to travel great distances for a one-off shot, a dedicated astrophotography site must be in close proximity to the owners/operators.
The third objective is to minimize the telescope's distance from any top-100 global engineering university.
The list of universities is provided courtesy of \cite{usnwr2024universities}, though most of the top-100 are located in the United States, China, UK, and Australia.

\subsubsection{Findings}

The final optimization problem for astrophotography is provided in Eq. (\ref{eq:astrophotography_optimization}).

\begin{equation} \label{eq:astrophotography_optimization}
    \begin{aligned}
        \min_{\phi,\lambda} \quad & M_{artificial}(\phi, \lambda)\\
        &\overline{O}(\phi, \lambda)\\
        &D_{university}(\phi, \lambda)\\
        \hfill\\
        \textrm{subject to} \quad & \phi \in \Phi_{land}\\
        &\lambda \in \Lambda_{land}\\
    \end{aligned}
\end{equation}

The optimization problem in Eq. (\ref{eq:astrophotography_optimization}) will be solved with both the repair and Gauss lattice-based crossover approaches.
For the Gauss lattice-based crossover approach, the number of quantiles in the lattice was $n_q = 10$, and the number of nodes at each quantile was $n_p = 12$.

If the proximity to owner/operators is ignored, Lake Victoria in Africa is the optimal location for astrophotography.
Lake Victoria enjoys virtually no artificial light pollution, and an average cloud cover of 19\% over the year.
Notably, much of the world lacks significant light pollution -- only regions around urban centers experience an appreciable amount of artificial light.

\begin{figure}[ht!]
    \centering
    \includegraphics[width=\textwidth]{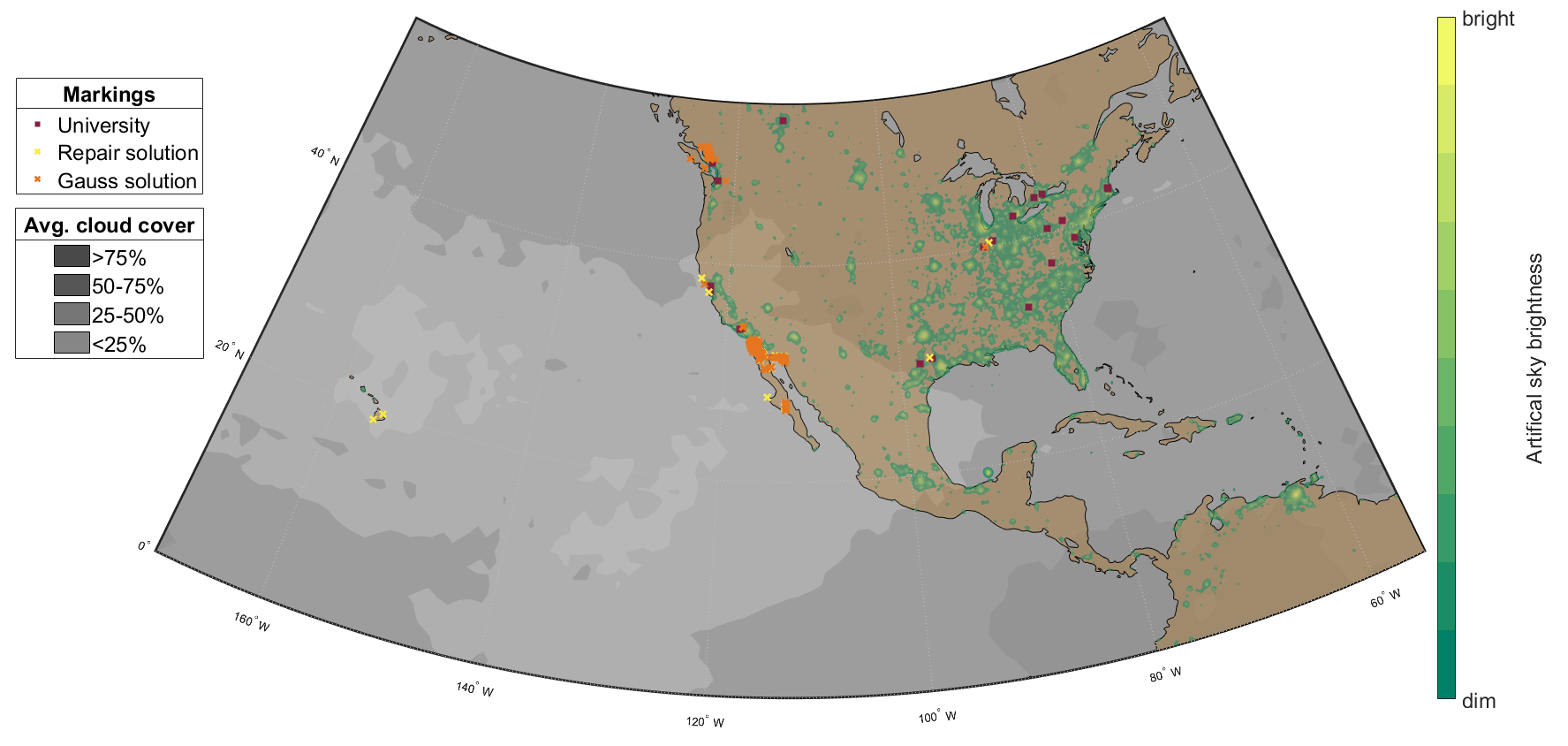}
    \caption{Gauss lattice Pareto front locations for astrophotography in the Americas.}
    \label{fig:gauss_pareto_americas_map}
\end{figure}

\begin{figure}[ht!]
    \centering
    \includegraphics[width=0.9\textwidth]{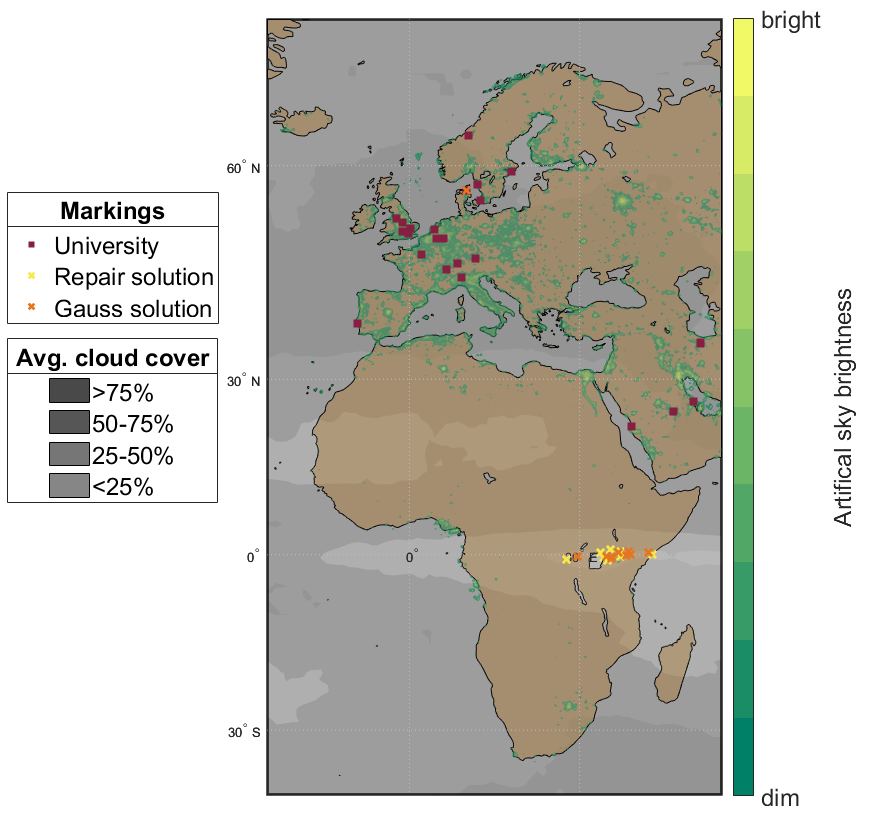}
    \caption{Gauss lattice Pareto front locations for astrophotography in the Europe and Africa.}
    \label{fig:gauss_pareto_euroafrica_map}
\end{figure}

\begin{figure}[ht!]
    \centering
    \includegraphics[width=0.9\textwidth]{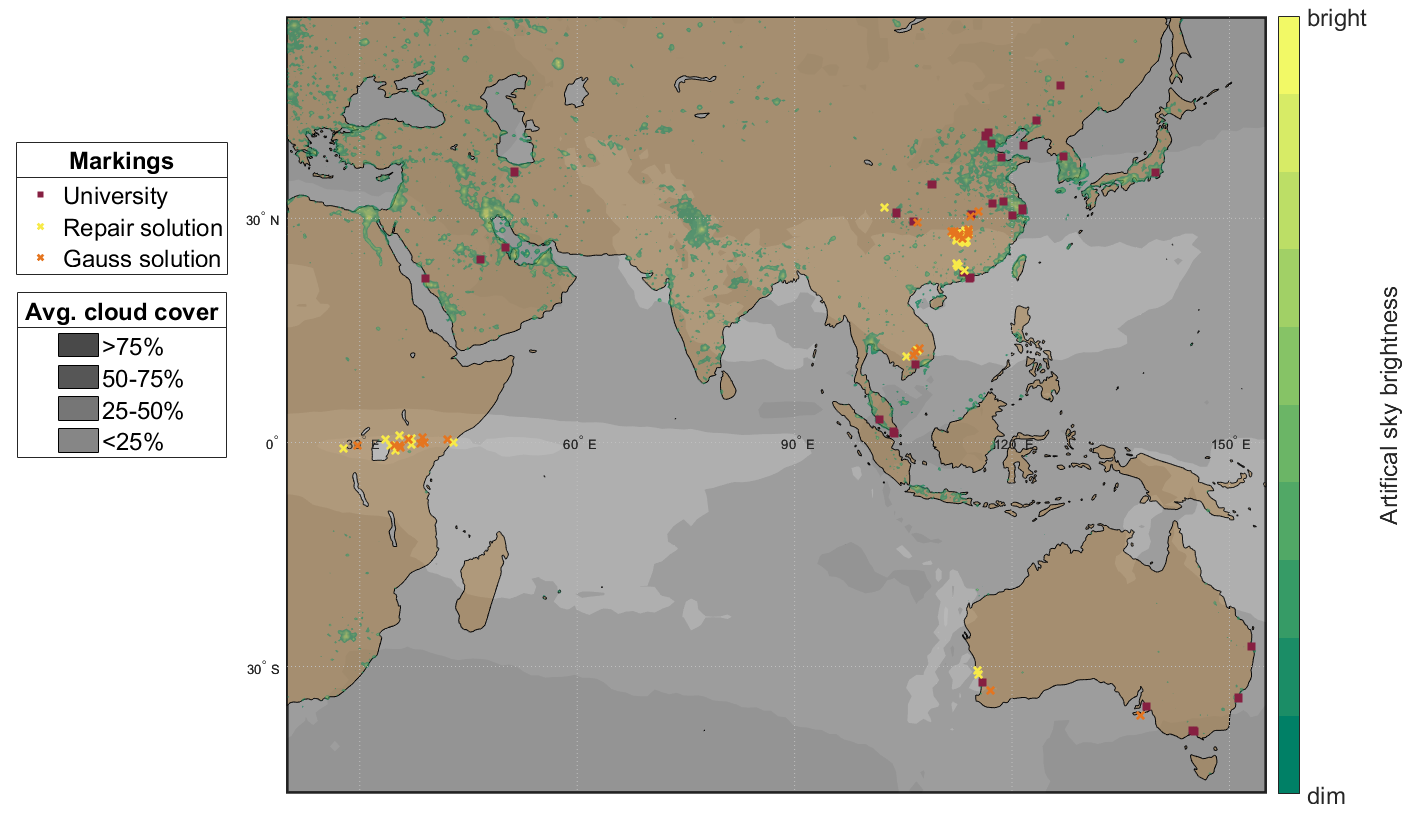}
    \caption{Gauss lattice Pareto front locations for astrophotography in Asia.}
    \label{fig:gauss_pareto_asia_map}
\end{figure}

A comparison of the Pareto front of optimal locations for both approaches is displayed in Figures \ref{fig:gauss_pareto_americas_map}, \ref{fig:gauss_pareto_euroafrica_map}, and \ref{fig:gauss_pareto_asia_map}.


The genetic algorithm converged in 1330 generations using the repair approach.
On the other hand, the genetic algorithm converged in 1133 generations using the Gauss lattice approach.
Both techniques resulted in points along the Pareto front that dominated points in the opposing front.
Parent A is said to dominate Parent B if the set of fitness values $\left(f_1, f_2,...,f_{N-1},f_N\right)$ satisfy the following conditions:

\begin{enumerate}
    \item $\left(f^A_{i} \leq f^B_{i}\right) \forall i$
    \item $\left(f^A_{1} < f^B_{1}\right) \vee \left(f^A_{2} < f^B_{2}\right)\vee ... \vee \left(f^A_{N-1} < f^B_{N-1}\right) \vee \left(f^A_{N} < f^B_{N}\right)$
\end{enumerate}

Regarding the Pareto front formed with the repair strategy, fifteen points are dominated by the Gauss lattice Pareto front.
Conversely, fourteen points on the Gauss lattice Pareto front are dominated by the repair Pareto front.

The interchanged domination between the two fronts is predominantly focused in two regions of the world.
Many of the Gauss lattice solutions in Southern California dominated Repair solutions in the same area.
Conversely, many repair solutions in China (particularly, the Guangdong province) dominated some Gauss lattice solutions in Southern California.
However, no Gauss lattice solution in Southern California that dominated a repair solution in Southern California was also dominated by a repair solution in China.
Due to the similar characteristics of the two regions, the apparent favoritism for exploration in Southern California by Gauss lattice versus the exploration of China by repair is ultimately a product of chance.
Solutions in these regions are balanced compromise between the three objectives -- universities are abundant, sky pollution is low outside the major cities, and average cloud cover is around 30\%.

Generally, both strategies led to the exploration of the same regions of the world.
However, Gauss lattice succeeded in finding nondominated solutions in the US Pacific Northwest, Denmark, and Southeastern Australia.
Conversely, repair succeeded in finding nondominated solutions on Hawaii.
Solution repair is a technique that is better posed to explore comparatively small feasible regions enclosed by large infeasible regions.
There are ways to make the lattice method more robust to such small feasible regions, such as solution preemption, or a hybrid lattice-repair strategy.

Overall, the final results are very similar between the two approaches.
Rigorous analysis of specific differences in the results is not particularly useful due to the stochasticity inherent in the procedure.
However, it is notable that the Gauss lattice strategy converged in 15\% fewer generations and produced a nearly equivalent Pareto front.
Faster convergence is an expected result, provided the data from Fig. \ref{fig:gens_convergence}.
Additionally, while not apparent in this particular instance, the Gauss lattice provides a higher confidence that the nondominated front resembles the true Pareto front, as demonstrated in Fig. \ref{fig:convergence}.
For engineering design problems, consistent solution quality is desirable.

\subsection{Applied orbit design sample problem}

Another constrained, continuous design problem of interest is orbit design.
All space missions, regardless of intent, require trajectory design.
A simple orbit design example is demonstrated for designing a set of complementary ground tracks, which may serve as the foundation for a number of use-cases.

An orbit set is sought that maximizes access time with a set of geographically distributed ground stations.
Some assumptions on the orbital dynamics are introduced to reduce the computational complexity of orbit propagation.
All relevant orbits are assumed to be circular, which disinvolves the argument of perigee and eccentricity elements.
The scenario commences from the satellite's ascending node, and the satellite is propagated for a week's time.
Under these assumptions, only the following orbital characteristics need to be specified:

\begin{itemize}
    \item Semi-major axis, $a$
    \item Inclination, $i$
    \item Longitude of Ascending Node, $\Omega$
\end{itemize}

The satellite's orbit period, $T$, is a direct function of its semi-major axis, defined in Eq. (\ref{eq:orbit_period}).

\begin{equation} \label{eq:orbit_period}
    T\left(a\right) = 2\pi\sqrt{\frac{\mu}{a^3}}
\end{equation}

The standard gravitational parameter, $\mu$, is $398600.435507 \unit{km/s^2}$ for an Earth orbit \cite{vallado_2013_astro}.
In this example problem, orbits of interest are restricted to inclinations below $60^{\circ}$.
As a simple approximation, orbit ground tracks are modeled as sinusoids.
In the sinusoid model, a satellite's latitude $\phi$ as a function of time is provided in Eq. (\ref{eq:sat_lat}).

\begin{equation} \label{eq:sat_lat}
    \phi(i,a,t) = i\sin\left(2\pi \frac{t}{T(a)}\right)
\end{equation}

Correspondingly, a satellite's longitude $\lambda$ as a function of time, in units of degrees, is provided in Eq. (\ref{eq:sat_lon}).

\begin{equation} \label{eq:sat_lon}
    \lambda(a,\Omega,t) = 2\pi\left(\frac{t}{T(a)} - \frac{t}{24}\right) + \Omega
\end{equation}

The factor $2\pi\left(\frac{t}{24}\right)$ in Eq. (\ref{eq:sat_lon}) accounts for the Earth's rotation, which approximately completes a full revolution every 24 hours.

\subsubsection{Orbit design objectives}

Two objectives are considered for the orbit design sample problem.
The first objective is the orbit access time with the designated ground stations.
In the scenario, there are three geographically distributed ground stations, detailed in Table \ref{tab:ground_stations}.

\begin{table}[!ht]
    \caption{ \label{tab:ground_stations} Ground stations for access}
    \centering
    \begin{tabular}{ c c c }
        \hline
        Ground Station & Latitude & Longitude \\
        \hline
        Blacksburg, USA & $37.226754^{\circ}$ & $-80.432546^{\circ}$ \\
        Geneva, CHE & $46.308158^{\circ}$ & $6.134166^{\circ}$ \\
        Winton, AUS & $-22.485683^{\circ}$ & $143.167884^{\circ}$ \\
        \hline
    \end{tabular}
\end{table}

\begin{figure}
    \centering
    \begin{tikzpicture}
        \draw (0pt, 0pt) circle (60pt);
        \filldraw[fill=black] (150pt, 0pt) circle (3pt);
        \draw[dotted] (150pt, 0pt) -- (24pt, 54.9913pt);
        \draw[dotted] (0pt, 0pt) -- (150pt, 0pt);
        \draw[dotted] (0pt, 0pt) -- (24pt, 54.9913pt);

        % Draw right angle
        \draw (22.8pt, 52.2414pt) -- (25.5495pt, 51.0414pt);
        \draw (25.5495pt, 51.0414pt) -- (26.7495pt, 53.7909pt);

        \draw (0pt, 0pt) -- (33.1845pt, 49.9879pt);
        \draw (0pt, 0pt) -- (60pt, 0pt);
        \draw (10pt,0pt) arc (0:56.4218:10pt);
        \draw[->] (16.0000pt, 36.6606pt) arc (66.4218:56.4218:40pt);
        \node[above] at (11.5000pt, 32.6606pt ) {\scriptsize $10\degree$};
        % \draw[->,>=stealth'] (8.3313pt, 5.5307pt) arc (33.5782:90:10pt);
        \node[above] at (13.5pt, 0pt) {$\psi$};

        \node[below] at (0, -60pt) {Earth};
        \node[right] at (152pt, 0pt) {Satellite};
    \end{tikzpicture}
    \caption{The resolvable arc of the Earth's surface from a satellite's perspective, resulting in the apparent disk radius, $\psi$. A $10\degree$ arc is removed from the limb because it is assumed to be unresolvable.}
    \label{fig:arclen_lim}
\end{figure}
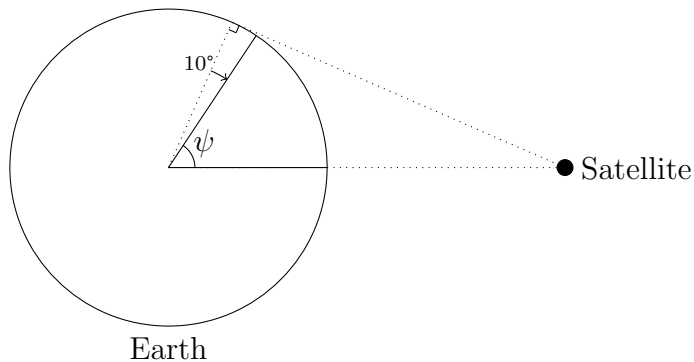

If the satellite's projected latitude-longitude position is within a limiting distance of a ground station, that ground station is considered accessible to the satellite.
The limiting distance is calculated based on the satellite's field of view.
In turn, the satellite's field of view is determined by its altitude.
The satellite's view projected onto the Earth's surface forms a disk, referred to hereafter as the apparent disk.
Depicted in Fig. \ref{fig:arclen_lim}, the radius of the apparent disk, $\psi$, is determined from arc of the Earth's surface resolvable from the satellite's position in space.

While the satellite's field of view geometrically extends to the Earth's limb, it is assumed the distant edge of the arc is non-resolvable.
Therefore, as a simple assumption, $10^{\circ}$ is subtracted from the arc to remove the non-resolvable region, yielding the field of view ``apparent disk''.
The size of the apparent disk, as a function of the satellite's altitude, $h$, is provided in Eq. (\ref{eq:apparent_disk}).

\begin{equation} \label{eq:apparent_disk}
    \psi = \cos^{-1}\left(\frac{R_{Earth}}{R_{Earth} + h}\right) - 10^{\circ}
\end{equation}

The satellite's altitude is directly related to its semi-major axis by $a = R_{Earth} + h$.

Notably, the apparent disk equation is an approximation, and the model can result in non-physical negative values of $\psi$ for low altitudes.
However, $\psi > 0$ for altitudes greater than 99\unit{km}.
All orbits considered in this analysis must de facto fly at an altitude above 99\unit{km}, so the limit $h > 99\unit{km}$ of the model in Eq. (\ref{eq:apparent_disk}) is acceptable within the given problem context.

Equation (\ref{eq:in_view}) describes how instantaneous access $\delta(t)$ is calculated for the $i^{th}$ ground station and $j^{th}$ satellite.
Note the use of the haversine formula, which was defined in Eq. (\ref{eq:haversine_formula}).

\begin{equation} \label{eq:in_view}
    \delta_{i,j}(t) = \begin{cases}
        1 & hav\left(\phi_i, \lambda_i, \phi_j(t), \lambda_j(t)\right) < \psi_j\\
        0 & otherwise \\
     \end{cases}
\end{equation}

Each orbit is propagated in discrete, 1 minute intervals for the duration of the scenario.
The total access time, denoted as $A_{total}$, is computed as the aggregate sum of time steps in which a ground stations is accessible.
At any time $t$ in the scenario, access at that time, denoted as $\delta(t)$, is a boolean value.
The total access time is defined in Eq. (\ref{eq:access_objective}).

\begin{equation} \label{eq:access_objective}
    A_{total} = \sum_{i=1}^{N_{stations}} \sum_{t=1}^{T} \left(\delta_{i,1}(t) \vee \delta_{i,2}(t) \vee ... \vee \delta_{i,N-1}(t) \vee \delta_{i,N}(t)\right)
\end{equation}

If normalized to a percentage, a total access time of $A_{total} = 1.0$ indicates that all three ground stations were accessible for the entirety of the scenario.

The second objective for orbit design is the expended energy, $\Delta{V}$, of each satellite in the orbit to reach its target altitude.
The energy to reach the desired orbit is only a function of the target orbit's semi-major axis, $a$.
In the context of optimization, the expended energy objective serves as a surrogate for monetary cost.
The $\Delta{V}$ is computed from the Hohmann transfer energy from a circular parking orbit.

It's assumed the parking orbit is already at the correct inclination from the launch.
The semi-major axis of the parking orbit, $a_{park}$, is $200\unit{km}$ altitude, such that $a_{park} = R_{Earth} + 200\unit{km}$.
The total $\Delta{V}$ incurred by the Hohmann transfer from the parking orbit is given in Eq. (\ref{eq:delta_v_objective}).

\begin{equation} \label{eq:delta_v_objective}
    \Delta{V}(a) = \sqrt{\frac{2\mu}{a_{park}} - \frac{\mu}{a_t}} -\sqrt{\frac{\mu}{a_{park}}} + \sqrt{\frac{\mu}{a}} - \sqrt{\frac{2\mu}{a} - \frac{\mu}{a_t}}
\end{equation}

The semi-major axis of the transfer orbit is $a_t = \frac{a_{park} + a}{2}$.

For multiple satellites on orbit, the total $\Delta{V}$ cost is the sum of individual $\Delta{V}$ costs for each satellite.
The cost objective is defined in Eq. (\ref{eq:total_delta_v_objective}).

\begin{equation}  \label{eq:total_delta_v_objective}
    \Delta{V}_{total} = \sum_{i=1}^{N}  \Delta{V}(a_i)
\end{equation}

A considerable consequence of using $\Delta{V}$ as a cost metric is that it aggregately penalizes high altitude orbits.
While a set of geosynchronous satellites may provide continuous access with the ground stations, it requires maximal cost.

\subsubsection{Orbit design constraints}

In the example presented here, constraints of practice are imposed, as opposed to physical constraints.
Most of near-Earth space is not used, or even considered, for common satellite operations.
The ESA's Annual Space Environment Report \cite{esa2023spaceenvironment} provides a detailed analysis of common orbit strata, and is leveraged here to determine some practical satellite orbit altitudes.
Specifically, the considered admissable bands of altitude are in Low Earth Orbit (LEO), Geosynchronous Earth Orbit (GEO), and some semi-synchronous orbits in Medium Earth Orbit (MEO), elaborated in Eq. (\ref{eq:orbit_bands}).
The selected semi-synchronous bands revolve around the Earth approximately 2, 3, and 4 times per day.

\begin{equation} \label{eq:orbit_bands}
    h_{feasible}, i_{feasible} = \begin{cases}
        350\unit{km} < h < \num{2000}\unit{km} & 45^{\circ} \leq i \leq 60^{\circ} \\
        \num{10185}\unit{km} < h < \num{10585}\unit{km} & 45^{\circ} < i < 60^{\circ} \\
        \num{13729}\unit{km} < h < \num{14129}\unit{km} & 45^{\circ} < i < 60^{\circ} \\
        \num{20032}\unit{km} < h < \num{20432}\unit{km} & 45^{\circ} < i < 60^{\circ} \\
        \num{35000}\unit{km} < h < \num{36500}\unit{km} & i \leq 15^{\circ} \\
    \end{cases}
\end{equation}

At the lowest admissable altitude (350\unit{km}), the diameter of the satellite's field of view, is $2\psi = 17.1\degree$ -- a relatively narrow field of view.
Considering many LEO satellites are nadir-locked, a small field of view for low-flying satellites is consistent with current practice.

\begin{figure}[ht!]
    \centering
    \includegraphics[width=0.6\textwidth]{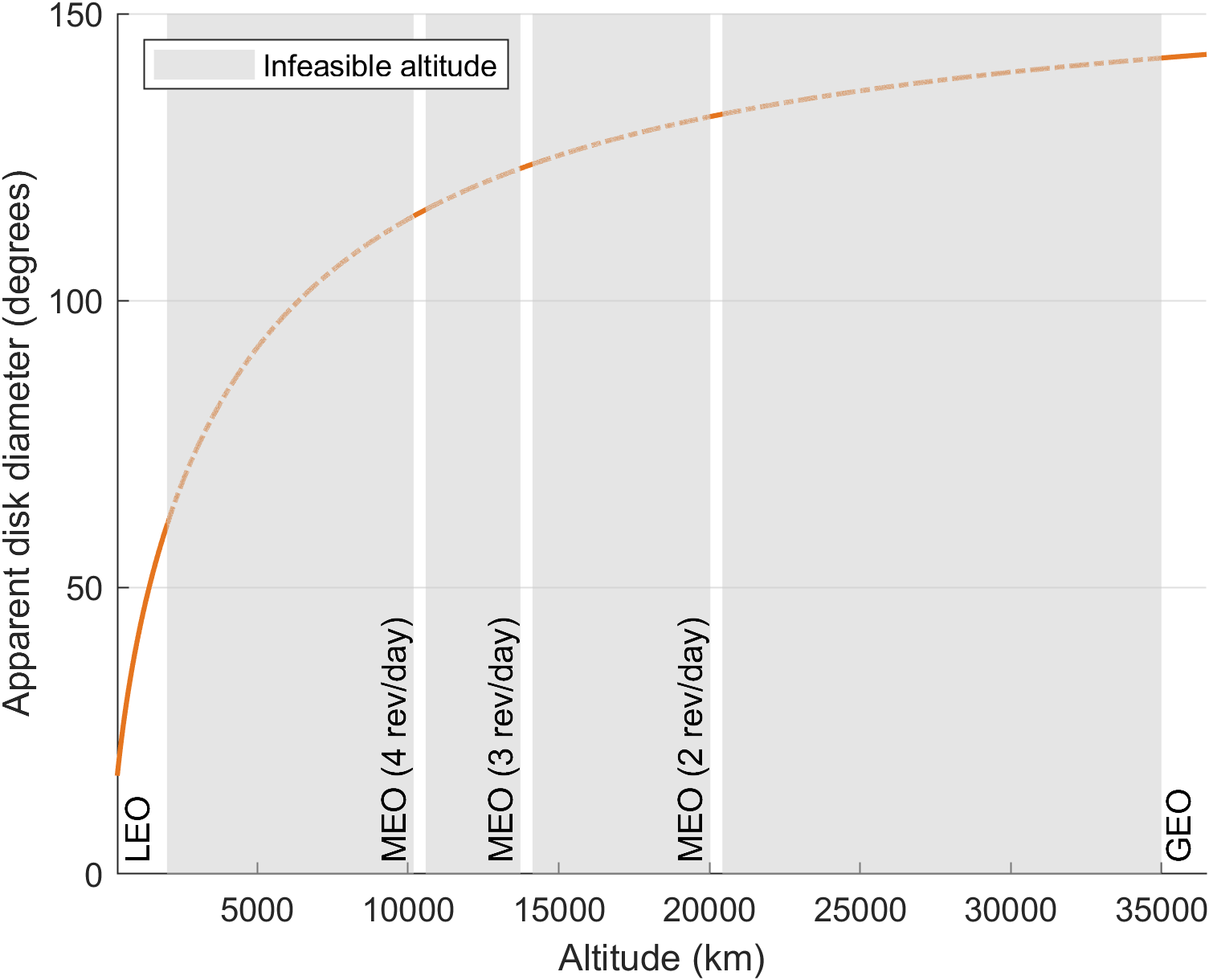}
    \caption{Size of the apparent disk with increasing altitude, with shaded infeasible regions.}
    \label{fig:apparent_disk_diameter}
\end{figure}

The diameter of the apparent disk, $2\psi$, is shown as a function of altitude in Fig. \ref{fig:apparent_disk_diameter}.
The gray regions in Fig. \ref{fig:apparent_disk_diameter} represent the invalid regions, whereas the white regions represent the valid orbit bands.
Notably, the semi-synchronous bands are small relative to the search space.
Custom lattice logic or a hybrid lattice-repair algorithm may be better posed to guide search in the design space.
However, the added complexity of such an algorithm is deemed out of scope for this work.

\subsubsection{Findings}

The final optimization problem for orbit design is provided in Eq. (\ref{eq:groundtrack_optimization}), where the orbits are optimized for access time and fuel expenditure while constrained to common practical orbit bands.

\begin{equation} \label{eq:groundtrack_optimization}
    \begin{aligned}
        \min_{\phi,\lambda} \quad &-A_{total}\left(t\right)\\
        & \Delta{v}_{total}\\
        \hfill\\
        \textrm{subject to} \quad & a \in a_{admissable}\\
        &i \in i_{admissable}\\
    \end{aligned}
\end{equation}

The optimization problem described in Eq. (\ref{eq:groundtrack_optimization}) will be solved with both the Gauss lattice-based crossover and penalty approaches.
In this optimization problem, the chromosome is variable length, so that a solution candidate may be composed of up to three satellites.
For the Gauss lattice-based crossover approach, the number of quantiles in the lattice was $n_q = 20$, and the number of nodes at each quantile was $n_p = 12$.
The quantile number $n_q$ was chosen strategically, so that the crossover between a 4 revolution per day MEO orbit and a 2 revolution per day MEO orbit would result in some lattice points in the 3 revolution per day MEO orbit band.

Regarding mutation for the lattice-based approach, the mutated values are selected from a precomputed set of one thousand admissable combinations of semi-major axis and inclination.
The majority of the values were randomly generated, but some were preempted -- more discussion on preemption is provided in Section \ref{subsec:mutation}.
The preempted values are provided in Table \ref{tab:preempted_alleles}.

\begin{table}[!ht]
    \caption{\label{tab:preempted_alleles} Preempted alleles for orbit design}
    \centering
    \begin{tabular}{ c c c }
        \hline
        Semi-major Axis (\unit{km}) & Inclination (\degree) & Comments \\
        \hline
        6828 & $53$ & Mock Starlink orbit \\
        16763 & $52$ & Semi-sync orbit (4 revs/day) \\
        20307 & $57$ & Semi-sync orbit (3 revs/day) \\
        26560 & $55$ & Mock GPS orbit  \\
        42164 & $0.0$ & Standard GEO orbit \\
        \hline
    \end{tabular}
\end{table}

Regarding the penalty approach, the ``Death Penalty'' is used in this example \cite{ponsich2008constraint}.
The death penalty is the simplest penalty method, as it requires no information about the domain.
In a problem as heavily constrained as this one, the penalty approach may struggle to find admissable regions.
As noted in \cite{yokoyama2005modified}, penalty methods in trajectory optimization problems are prone to stagnation and premature convergence.
Therefore, the penalty method will be run a total of five times, and the convergence criterion will be set so that 50 succeeding generations must elapse without improvement to converge.

\begin{figure}[ht!]
    \centering
    \includegraphics[width=0.7\textwidth]{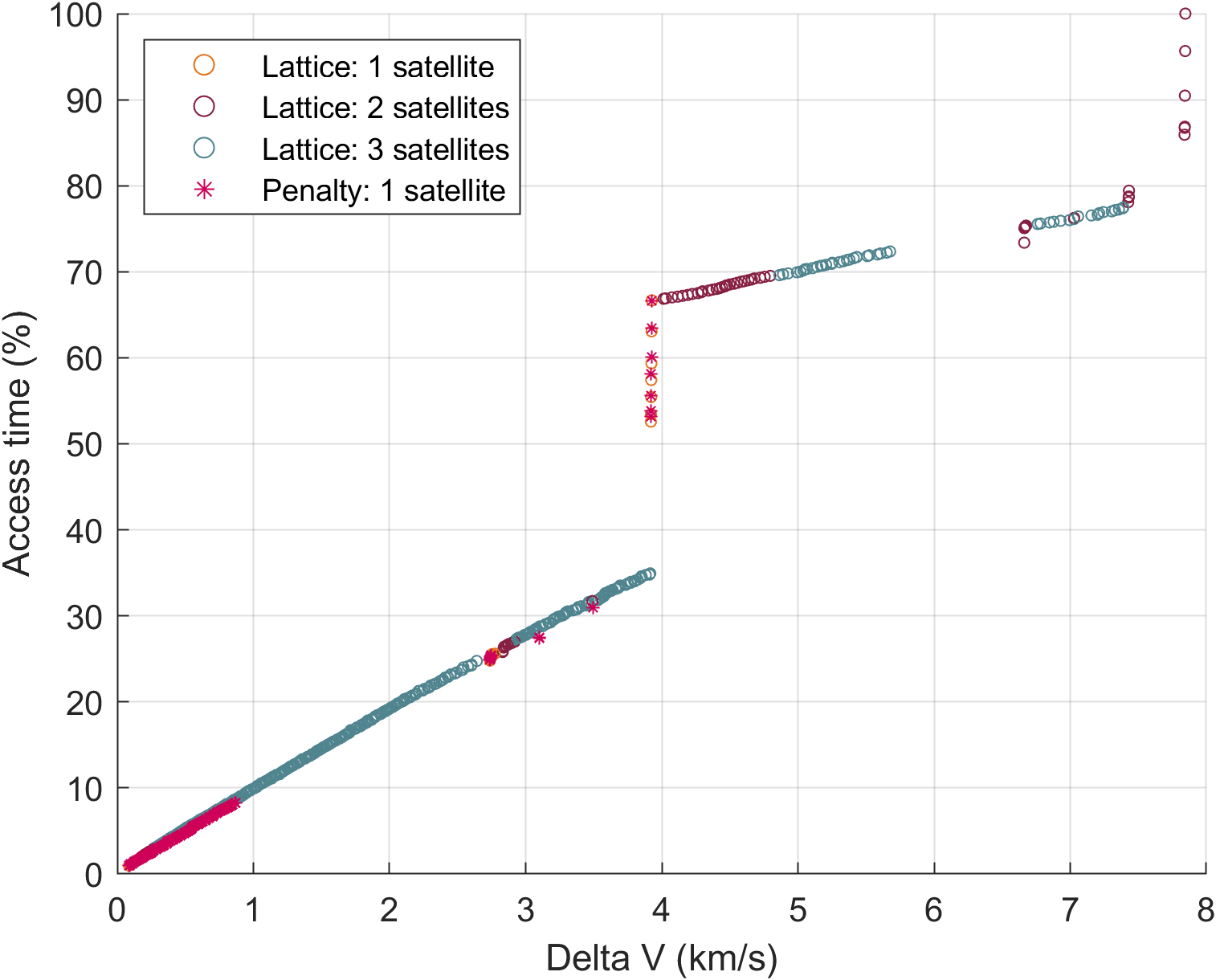}
    \caption{Pareto fronts for orbit solutions from both the lattice and penalty methods, with the fuel cost objective on the X-axis, and the ground station access objective on the Y-axis.}
    \label{fig:groundtrack_pareto_front}
\end{figure}

Figure \ref{fig:groundtrack_pareto_front} compares the Pareto fronts returned by the different approaches.
The solutions from the Gauss lattice-based approach were produced from a single run, while the solutions from the penalty approach are the nondominated union of solutions from all five runs.
The Gauss lattice technique converged in about 20,000 generations.
Each of the five runs using the penalty approach converged in about 2,500 generations due to stagnation.

The penalty method results failed to include any orbit designs containing more than one satellite orbit with comparatively brief access time.
This result is not surprising, given that about 12\% of the selectable altitude domain is admissable.
Independently, an admissable inclination must also be selected, and the resulting orbit must provide competitive value for its cost.
The likelihood of such an event occurring in the framework of a traditional genetic algorithm is vanishingly small.
The lattice approach, on the other hand, yielded many multi-orbit solutions with more robust ground station access.

% The penalty method yielded only single orbit solutions with low access time, whereas the lattice method yielded many multi-orbit solutions with more robust ground station access.

Regarding single satellite solutions, the results of the lattice approach were nearly identical to results of the penalty approach.
The low cost ($\Delta V < 2 \unit{km/s}$), low access single satellite solutions from either approach are entirely LEO orbits.
Intuitively, more satellites provide more robust access, which is why multi-satellite solutions dominate the Pareto front.
As illustrated in Fig. \ref{fig:apparent_disk_diameter}, relatively small increases in altitude yield substantial increases in field of view, explaining the linear trend in the left half of the Pareto front.

The near-vertical jump around $\Delta V = 4 \unit{km/s}$ is the cost threshold where GEO satellites can be considered.
Notably, both the lattice and penalty approaches found single satellite GEO solutions.
In all cases, the single GEO solutions were located at a longitude above the Atlantic ocean, where both the Blacksburg and Geneva stations are accessible from GEO altitude.
The 2- and 3-satellite solutions just beyond the GEO threshold are solutions comprised of one GEO and one or two LEO satellites.

\begin{figure}[ht!]
    \centering
    \includegraphics[width=0.7\textwidth]{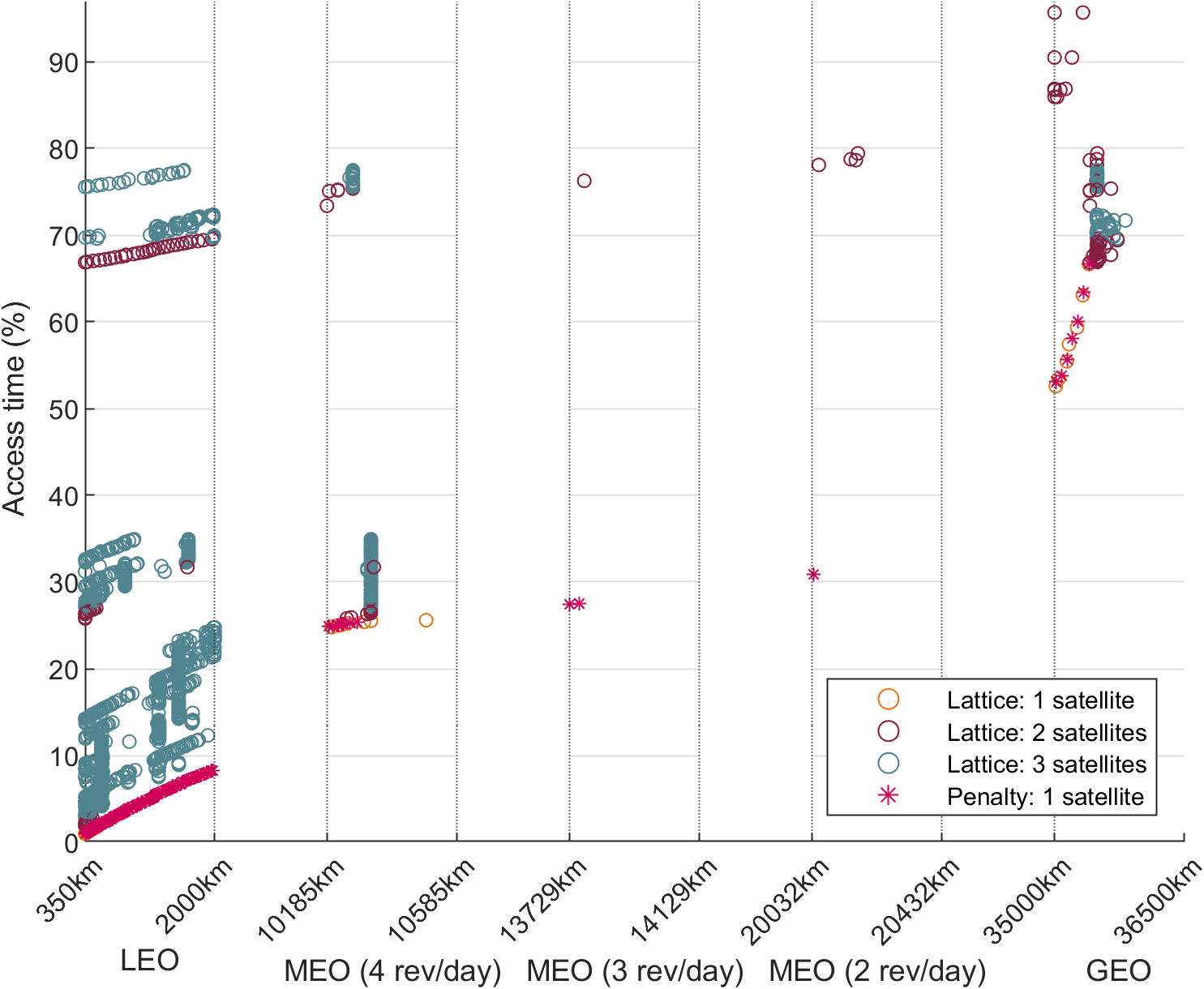}
    \caption{Altitudes of the satellite orbits from optimal solutions produced by Gauss lattice and penalty methods.}
    \label{fig:groundtrack_pareto_front_altitudes}
\end{figure}

Figure \ref{fig:groundtrack_pareto_front_altitudes} visualizes the altitudes of the satellite solutions from Fig. \ref{fig:groundtrack_pareto_front}.
Examining Fig. \ref{fig:groundtrack_pareto_front_altitudes}, the distinct difference between the lattice and repair solutions is that the repair approach found multi-satellite mixed LEO-MEO solutions that achieved equivalent access to single satellite MEO solutions.
Figure \ref{fig:groundtrack_pareto_front} shows that these mixed orbit regime solutions were more cost-efficient than the single satellite solutions.

\begin{figure}[ht!]
    \centering
    \includegraphics[width=0.7\textwidth]{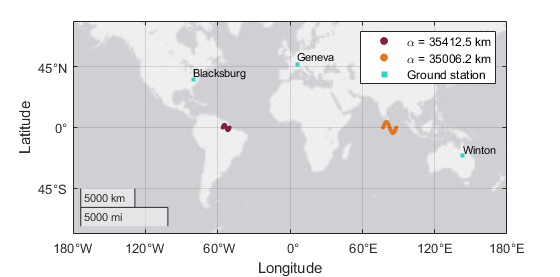}
    \caption{Solution with 100\% access, $\Delta V = 7.85 \unit{km/s}$.}
    \label{fig:gauss_groundtrack_sol_281}
\end{figure}

The solutions which achieved the highest access were comprised of two GEO satellites.
For reference, the fuel cost of launching three satellites to GEO is $\Delta{V} = 11.796 \unit{km/s}$.
However, the lattice approach uncovered a solution with only two GEO satellites that provided uninterrupted coverage of all ground stations.
The orbits straddled two longitudes between the ground stations.
The ground tracks of the solution are shown in Fig. \ref{fig:gauss_groundtrack_sol_281}.

At GEO altitude, the satellites' field of view have a limiting distance of approximately $\psi = 71\degree$.
The Blacksburg and Geneva stations are accessible to the westward satellite, and the Winton station is accessible to the eastward satellite.
Notably, both sit at an altitude just below true geosynchronous orbit, which is an altitude of $h_{GEO} = 35786\unit{km}$.
The result is that both satellites drift eastward over time.
Because the propagation elapses only a week, the altitudes and initial longitudes were selected such that the drift over the course of the week is not enough to remove the ground stations from view.
The lower altitudes result in a slightly lower ${\Delta}V$, so this solution dominates a solution of true geosynchronous satellites within the bounds of the problem setup.
If indefinite access is the true design goal for the orbit, the satellites' altitudes need only be increased to $35786\unit{km}$, at commensurate cost.

\begin{figure}[ht!]
    \centering
    \includegraphics[width=0.7\textwidth]{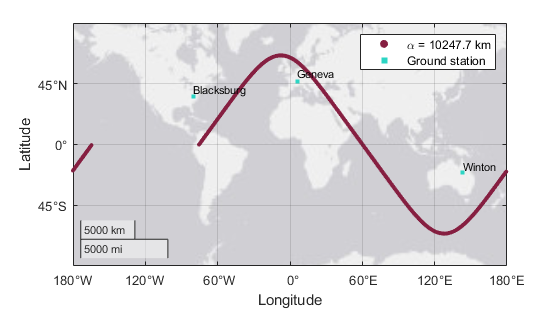}
    \caption{Solution with 25\% access, $\Delta V = 2.74\unit{km/s}$.}
    \label{fig:gauss_groundtrack_sol_44}
\end{figure}

Another region of interest on the Pareto front is the regime single-satellite solutions that achieved 25\% coverage.
All solutions in this regime featured a $60\degree$ inclination semi-synchronous MEO orbit that completed approximately four revolutions per day.
Initial latitude varied among solutions, but a representative example is shown in Fig. \ref{fig:gauss_groundtrack_sol_44}.

Notably, at $60\degree$ inclination, the ground track travels about $15\degree$ north of the Geneva and $35\degree$ south of Winton.
The altitude of this orbit provides a limiting distance of $\psi = 57\degree$.
At higher inclination, the Geneva and Winton ground stations are accessible for a longer duration.

Overall, the lattice approach succeeded in finding many more diverse solutions along the Pareto front in one run than the death penalty approach achieved in five runs.

\section{Conclusion}\label{sec:conclusion}

% The main conclusions of the study may be presented in a short Conclusions section, which may stand alone or form a subsection of a Discussion or Results and Discussion section.

% In conclusion, this work presented a novel lattice-based crossover methodology for GAs that effectively handles multidimensional constraints for continuous decision variables.
In conclusion, GAs are a powerful optimization tool for complex engineering design problems.
This work presented a novel lattice-based crossover methodology for GAs that is effective for handling nonlinear multidimensional constraints.
The lattice-based approach achieved better performance compared to traditional repair and penalty methods, with regards to both solution convergence and population diversity.
The uniform lattice and Gaussian lattice structures were both demonstrated, with the Gaussian lattice approach showing considerable utility for geographical search and orbit design.
The lattice-based method is extensible to arbitrary lattice structures, permitting configurability for other domain-specific approaches.
Additionally, a feature for solution preemption was proposed within the mutation operator for the lattice-based GA, which encourages exploration in statistically unlikely regions of the search space.

A Monte Carlo analysis and accompanying example applications were provided to demonstrate the lattice-based GA.
Specific takeaways are:
\begin{itemize}
    \item Monte Carlo simulations demonstrated that the lattice-based method finds solutions two orders of magnitude closer to the optima in fewer generations.
    \item In the geographic search problem for optimal telescope placement, the lattice-based GA converged to the Pareto front 15\% faster than traditional methods.
    \item For the satellite constellation design problem, the lattice-based method discovered an order of magnitude more Pareto-optimal solutions within a highly constrained space.
\end{itemize}

Overall, it was demonstrated that the lattice-based GA possesses superior exploration capabilities for optimization problems with constraints or epistatic decision variables.
The lattice-based approach facilitates a more comprehensive traversal of the solution space, leading to faster convergence and a higher probability of finding high-quality solutions in complex, multi-objective optimization problems.

Future research directions include investigating alternative lattice structures and the potential of hybrid lattice-repair techniques.
This work opens avenues for further development and application of lattice-based methods within the broader field of evolutionary computation.

%%-------------------------------------------------------------

%% Bibliography

\newpage
\bibliographystyle{elsarticle-num}
\bibliography{cco}
% \bibliography{cco}

%%-------------------------------------------------------------

%% Nomenclature & Acronyms

% \newpage
% \printacronyms[]

% Tells LaTeX to generate nomenclature output file
% \subfile{meta/nomeclature}

%%-------------------------------------------------------------

\end{document}